\documentclass[journal,transmag]{IEEEtran}

\usepackage{multicol}
\usepackage{amsmath}

\usepackage{lineno,hyperref}
\usepackage{subcaption}
\usepackage{booktabs}
\usepackage{multirow}
\usepackage{arydshln}
\usepackage{enumitem}

\usepackage{lineno,hyperref}
\usepackage{multirow}
\usepackage{graphicx}
\usepackage{subcaption}
\usepackage{mathtools}
\usepackage[export]{adjustbox}

\usepackage{setspace}
\usepackage{wrapfig,booktabs}

\usepackage{amsmath}
\usepackage[shortcuts, acronym]{glossaries}
\makeglossaries

\newacronym{eum}{EUM}{Embedding Unmasking Model}
\newacronym{fnmr}{FNMR}{false non-match rate}
\newacronym{fmr}{FMR}{false match rate}
\newacronym{srt}{SRT}{Self-restrained Triplet Loss}
\newacronym{nist}{NIST}{National Institute of Standards and Technology}
\newacronym{eer}{EER}{Equal Error Rate}
\usepackage[dvipsnames,table,xcdraw]{xcolor}
\definecolor{revision}{rgb}{0,0,0}
\definecolor{2revision}{RGB}{0,0,0}

\ifCLASSINFOpdf

\else

\fi

\begin{document}
\title{Self-restrained Triplet Loss for Accurate Masked Face Recognition}

\author{Fadi Boutros$^{a,b}$, Naser Damer$^{a,b,*}$,  Florian Kirchbuchner$^{a}$, Arjan Kuijper$^{a,b}$\\
$^{a}$Fraunhofer Institute for Computer Graphics Research IGD, Darmstadt, Germany\\
$^{b}$Mathematical and Applied Visual Computing, TU Darmstadt,
Darmstadt, Germany\\
Email: fadi.boutros@igd.fraunhofer.de
}

\maketitle

\begin{abstract}
Using the face as a biometric identity trait is motivated by the contactless nature of the capture process and the high accuracy of the recognition algorithms. After the current COVID-19 pandemic, wearing a face mask has been imposed in public places to keep the pandemic under control.
However, face occlusion due to wearing a mask presents an emerging challenge for face recognition systems. In this paper, we present a solution to improve masked face recognition performance. Specifically, we propose the Embedding Unmasking Model (EUM) operated on top of existing face recognition models. We also propose a novel loss function, the Self-restrained Triplet (SRT), which enabled the EUM to produce embeddings similar to these of unmasked faces of the same identities.
The achieved evaluation results on three face recognition models, two real masked datasets, and two synthetically generated masked face datasets proved that our proposed approach significantly improves the performance in most experimental settings. \footnote{
The SRT implementation, training and evaluation codes, pretrained models and the list of mask types and colors applied on IJB-C and LFW are publicly released for reproducibility of the result \url{https://github.com/fdbtrs/Self-restrained-Triplet-Loss}}.
\end{abstract}





\section{Introduction}
Face recognition is one of the preferable biometric recognition solutions due to its contactless nature and the high accuracy achieved by face recognition algorithms. 
Face recognition systems have been widely deployed in many application scenarios such as automated border control, surveillance, as well as convenience applications \cite{gorodnichy2014automated,lovisotto2017mobile,ARANDJELOVIC2018388}. However, these systems are mostly designed to operate on none occluded faces.
After the current COVID-19 pandemic,  wearing a protective face mask has been imposed in public places by many governments to reduce the rate of COVID-19 spread. This new situation raises a serious unusually challenge for the current face recognition systems.
Recently, several studies have evaluated the effect of wearing a face mask on face recognition accuracy \cite{DBLP:conf/biosig/DamerGCBKK20,DHS-Rally-2020,ngan2020ongoing,ngan2020ongoingpost}. 
These studies have reported the negative impact of masked faces on face recognition performance.
The main conclusion of these studies \cite{DBLP:conf/biosig/DamerGCBKK20,DHS-Rally-2020,ngan2020ongoing,ngan2020ongoingpost} is that the accuracy of face recognition algorithm with a masked face is significantly degraded, in comparison to unmasked face.

Motivated by this new circumstance we propose in this paper a new approach to reduce the negative impact of wearing a facial mask on face recognition performance. The presented solution is designed to operate on top of existing face recognition models and thus, avoid retraining existing solutions developed for unmasked face recognition. 
Recent works either proposed to train face recognition models with simulated masked faces \cite{anwar2020masked} or to train a model to learn the periocular area of the face images exclusively \cite{li2021cropping}.
Unlike these, our proposed solution does not require any modification or training of the existing face recognition model.
We achieved this goal by proposing the \gls{eum} operated on the embedding space. The input for \gls{eum} is feature embedding extracted from the masked face, and its output is new feature embedding similar to an embedding of an unmasked face of the same identity, whereas, it is dissimilar from any other embedding of other identities. 
To achieve that through our \gls{eum}, we propose a novel loss function, \gls{srt} to guide the \gls{eum} during the training phase.
The \gls{srt} shares the same learning objective with the triplet loss i.e. it enables the model to minimize the distance between genuine pairs and maximize the distance between imposter pairs.   Nonetheless, unlike triplet loss, the \gls{srt} can dynamically self-adjust its learning objective by focusing on minimizing the distance between the genuine pairs when the distance between the imposter pairs is deemed to be sufficient. 

The presented approach is evaluated on top of three face recognition models, ResNet-100 \cite{DBLP:conf/cvpr/HeZRS16}, ResNet-50 \cite{DBLP:conf/cvpr/HeZRS16} and MobileFaceNet \cite{DBLP:journals/corr/abs-1804-07573} trained with the loss function, Arcface loss \cite{DBLP:conf/cvpr/DengGXZ19}, to validate the feasibility of adopting our solution on top of different deep neural network architectures.
With a detailed evaluation of the proposed \gls{eum} and \gls{srt},
we reported the verification performance gain by the proposed approach on two real masked face datasets \cite{anwar2020masked,DBLP:conf/biosig/DamerGCBKK20} and two synthetically generated masked face datasets. 
We further experimentally supported our theoretical motivation behind our \gls{srt} loss by comparing its performance with the conventional triplet loss. The overall verification result showed that our proposed approach improved the performance in most of the experimental settings. For example, when the probes are masked, the achieved FMR100 measures  (the lowest \gls{fnmr} for \gls{fmr} $\leq$ 1.0 \%) by our approach on top of MobileFaceNet are reduced by $ \sim $ 28\% and 26\% on the two real masked face evaluation datasets.


In the rest of the paper, we discuss first the related works focusing on masked face recognition in Section \ref{sec:related_work}.
Then, we present our proposed \gls{eum} architecture and our \gls{srt} loss in  Section \ref{sec:mthodology}.
In Section \ref{sec:exp}, we present the experimental setups and implementation details applied in this work. 
Section \ref{sec:result} presents and discuss the achieved results.
Finally, a set of conclusions are drawn in Section \ref{sec:conclusion}.

\section{Related Work}  
\label{sec:related_work}
In recent years, significant progress has been made to improve face recognition verification performance with essentially non-occluded faces.
Several previous works \cite{DBLP:conf/iccv/SongGLLL19,DBLP:journals/corr/OpitzWPPB16,JEEVAN2022108308,DBLP:journals/pami/YangLQTZX17} addressed general face occlusion e.g. wearing sunglasses or a scarf. Nonetheless, they did not directly address facial mask occlusion (before the current COVID-19 situation).

After the current COVID-19 situation, four major studies evaluated the effect of wearing a facial mask on face recognition performance \cite{DBLP:conf/biosig/DamerGCBKK20,DHS-Rally-2020,ngan2020ongoing,ngan2020ongoingpost}. 
The National Institute of Standards and Technology (NIST) has published two specific studies on the effect of masked faces on the performance of face recognition solutions submitted by vendors using pre-COVID-19 \cite{ngan2020ongoing}  and post-COVID-19 \cite{ngan2020ongoingpost} algorithms. These studies are part of the ongoing Face Recognition Vendor Test (FRVT). The studies by the NIST concluded that wearing a face mask has a negative effect on face recognition performance. However, the evaluation by NIST is conducted using synthetically generated masks, which may not fully reflect the actual effect of wearing a protective face mask on the face recognition performance.
The recent study by Damer et al. \cite{DBLP:conf/biosig/DamerGCBKK20} has tackled this limitation by evaluating the effect of wearing a mask on two academic face recognition algorithms and one commercial solution using a specific collected dataset for this purpose from 24 participants over three collaborative sessions. The study indicates the significant effect of wearing a face mask on face recognition performance. A similar study was carried out by the Department of Homeland Security (DHS) \cite{DHS-Rally-2020}. In this study, several face recognition systems (using six face acquisition systems and 10 matching algorithms) were evaluated on a specifically collected dataset of 582 individuals. The main conclusion from this study is that the accuracy of most best-performing face recognition systems is degraded from 100\% to 96\% when the subject is wearing a facial mask.

Li et al. \cite{li2021cropping} proposed to use an attention-based method to train a face recognition model to learn from the periocular area of masked faces. The presented method showed improvement in the masked face recognition performance. However, the proposed approach is only tested on simulated masked face datasets, and it essentially only maps the problem into a periocular recognition problem.
A recent preprint by \cite{anwar2020masked} presented a small dataset of 269 unmasked and masked face images of 53 identities crawled from the internet. The work proposed to fine-tune FaceNet model \cite{DBLP:journals/corr/SchroffKP15} using simulated masked face images to improve the recognition accuracy. However, the proposed solution is only tested using a small dataset (269 images).
Recently, a rapid number of researches are published to address the detection of wearing a face mask \cite{loey2021hybrid,qin2020identifying}. These studies did not directly address the effect of wearing a mask on face recognition performance or presenting a solution to improve masked face recognition. 

Motivated by the recent evaluations efforts on the negative effect of wearing a facial mask on the face recognition performance \cite{DBLP:conf/biosig/DamerGCBKK20,DHS-Rally-2020,ngan2020ongoing,ngan2020ongoingpost} and driven by the need for exclusively developing an effective solution to improve masked face recognition, we present in this work a novel approach to improve masked face recognition performance. The proposed solution is designed to run on top of existing face recognition models. Thus, it does not require any retraining of the existing face recognition models as presented in next Section \ref{sec:mthodology}.

\section{Methodology} 
\label{sec:mthodology}
\begin{figure*}[ht!]
    \centering
    \includegraphics[width=0.75\textwidth]{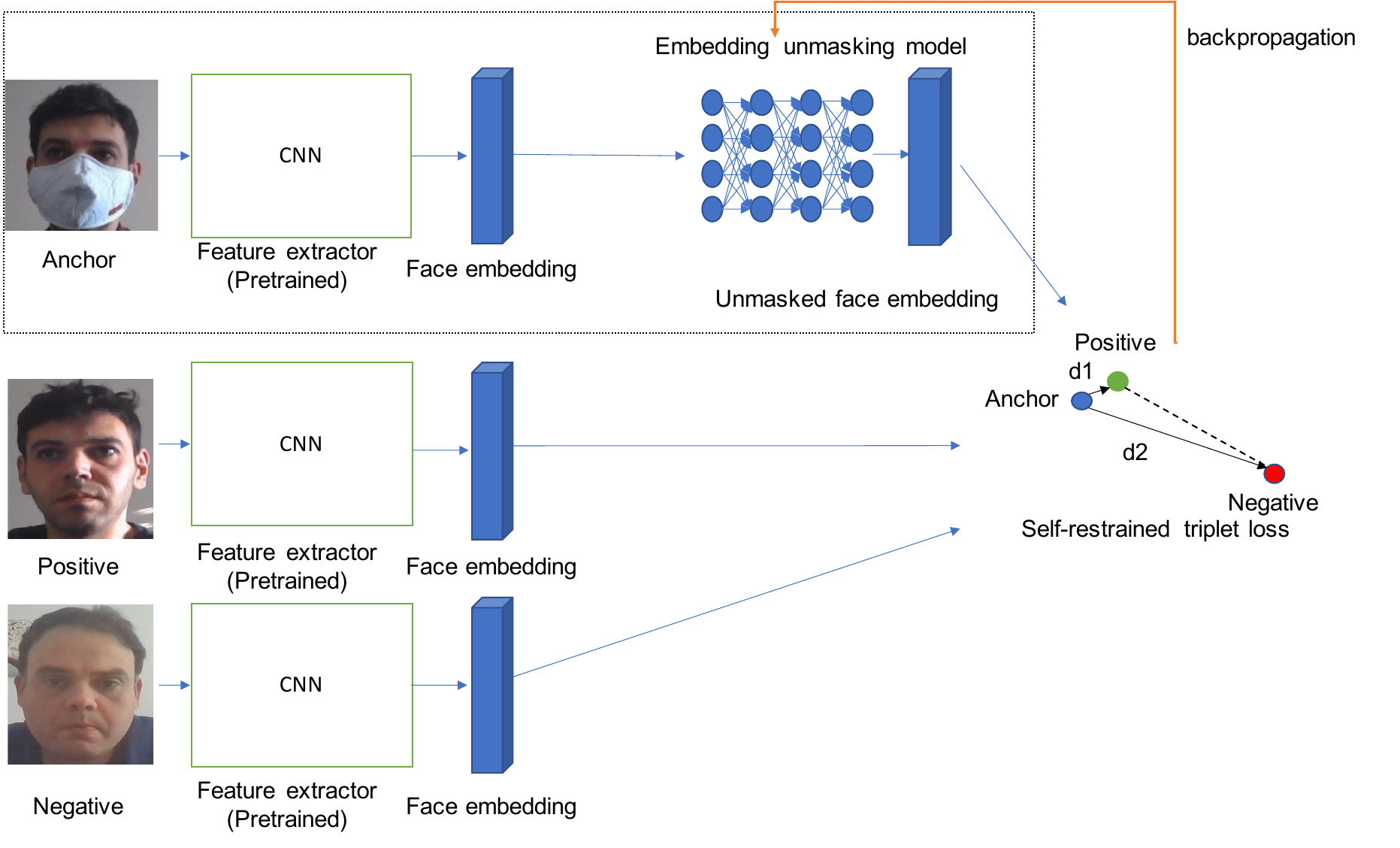}
    \caption{The workflow of the face recognition model with our proposed \gls{eum}. The top part of the figure (inside the dashed rectangle) shows the proposed solution in operation mode.
    Given an embedding obtained from the masked face, the proposed model is trained with \gls{srt} loss to output a new embedding that is similar to the one of the unmasked face of the same identity and different from the unmasked face embedding from all other identities.   }
    \label{fig:workflow}
\end{figure*}

In this section, we present our proposed approach to improve the verification performance of masked face recognition. The proposed solution is designed to operate on top of existing face recognition models.
To achieve this goal, we propose an \gls{eum}.
The input to our proposed model is a face embedding extracted from a masked face image, and the output is a so-called "unmasked face embedding", which is intended to be more similar to the embedding of the same identity without wearing a mask. Therefore, the proposed solution does not require any modification or training of the existing face recognition solution. 
Figure \ref{fig:workflow} shows an overview of the workflow of the proposed approach in training and operational modes. 

Furthermore, we propose the \gls{srt} to control the model during the training phase. Similar to the well-known triplet-based learning, the \gls{srt} loss has two learning objectives: 1) Minimizing the \textcolor{2revision}{intra-class} variation, i.e., minimizing the distance between genuine pairs of unmasked and masked face embeddings. 2) Maximizing the \textcolor{2revision}{inter-class} variation, i.e., maximizing the distance between imposter pairs of masked face embeddings. However, unlike the traditional triplet loss, the proposed \gls{srt} loss function can self-adjust its learning objective by only focusing on optimizing the \textcolor{2revision}{intra-class} variation when the \textcolor{2revision}{inter-class} variation \textcolor{revision}{ is deemed sufficient.} When the gap in \textcolor{2revision}{inter-class} variation is violated, our proposed loss behaves like a conventional triple loss.
The theoretical motivation behind our \gls{srt}-loss is presented along with the functional formulation later in this section.
In the following, this section presents our proposed \gls{eum} architecture and  the \gls{srt} loss.

\subsection{Embedding Unmasking Model Architecture}
The  \gls{eum} architecture is based on a fully connected neural network (FCNN). 
Having an FCNN architecture, where all neurons are connected in two consecutive layers, we can demonstrate a generalized EUM design. This is the case because this structure can be easily adapted to different input shapes, and thus can be adapted on the top of different face recognition models, motivating our decision to use FCNN.
The model input is a masked feature embedding (i.e., resulting from a masked face image) of size $D$ ($D$ depends on the face recognition network used), and the model output is a feature vector of the same size $D$. The proposed model consists of four fully connected layers (FC): an input layer, two hidden layers, and an output layer. The input size for all FC layers is of size $d$.
Each of the input and the hidden layers is followed by batch normalization (BN) \cite{DBLP:conf/icml/IoffeS15} and Leaky ReLU non-linearity activation function \cite{Maas2013RectifierNI}. The last FC layer is followed by BN.

\subsection{Unmasked Face Embedding Learning}
The learning objective of our model is to reduce the \gls{fnmr} of genuine unmasked-masked pairs.
The main motivation behind this learning goal is inspired by the latest reports on evaluating the effect of the masked faces on face recognition performance by the National Institute of Standards and Technology (NIST) \cite{ngan2020ongoing} and the recent work by Damer et al. \cite{DBLP:conf/biosig/DamerGCBKK20}. The NIST report \cite{ngan2020ongoing} stated that the false non-match rates (FNMR) are increased in all evaluated algorithms when the probes are masked.    
For the most accurate algorithms, the FNMR increased from 0.3\%  to 5\% at \gls{fmr} of 0.001\% when the probes are masked. On the other hand, the NIST report concluded that FMR appeared to be less affected when probes are masked.
A similar observation comes from the study by Damer et al. \cite{DBLP:conf/biosig/DamerGCBKK20}. This work reported that the genuine score distributions are significantly affected by masked probes \cite{DBLP:conf/biosig/DamerGCBKK20}. The study also reported that the genuine score distribution strongly shifts towards the imposter score distributions. On the other hand, the imposter score distributions do not seem to be strongly affected by masked face probes. 
One of the main observations of the previous studies in \cite{DBLP:conf/biosig/DamerGCBKK20,ngan2020ongoing}, is that wearing a face mask significantly increase the \gls{fnmr}, whereas the \gls{fmr} seem to be less affected by wearing a mask. Similar remarks have been also reported in our result (see Section \ref{sec:result}). Based on these observations, we motivate our proposed \gls{srt} loss function to focus on increasing the similarity between genuine pairs of unmasked and masked face embeddings, while maintaining the imposter distance at an acceptable level. In the following, we briefly present the naive triplet loss followed by our proposed \gls{srt} loss.

\begin{figure*}[ht]
     \centering
        \begin{subfigure}[b]{0.30\textwidth}
         \centering
         \includegraphics[width=\textwidth]{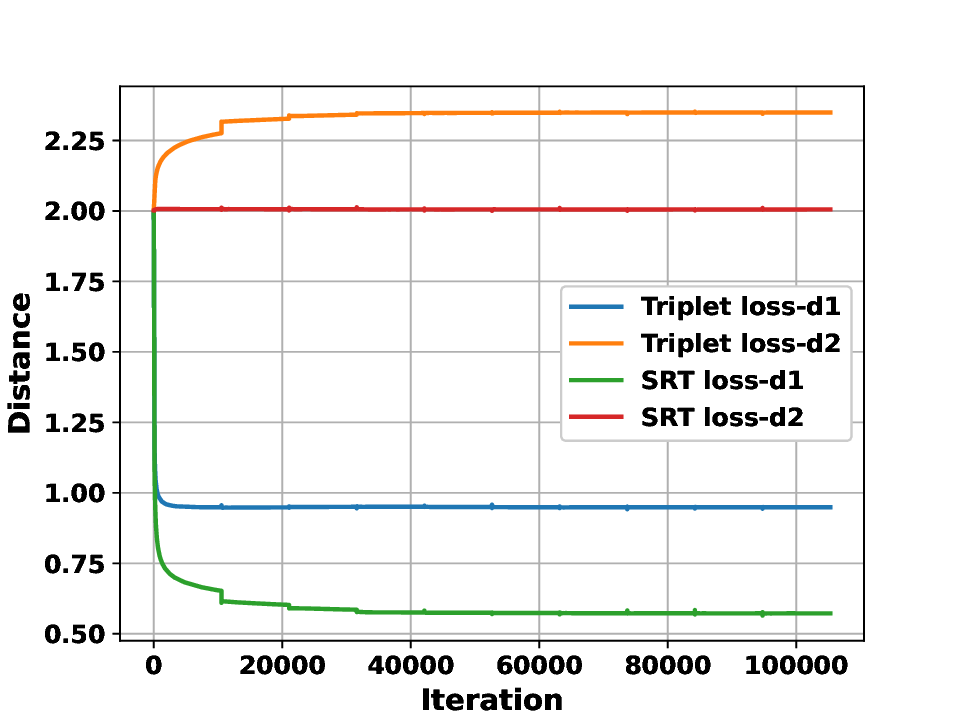}
         \caption{\textcolor{revision}{ ResNet-100}}
         \label{fig:triple_vs_srt_resnet100}
     \end{subfigure}
     \begin{subfigure}[b]{0.30\textwidth}
         \centering
         \includegraphics[width=\textwidth]{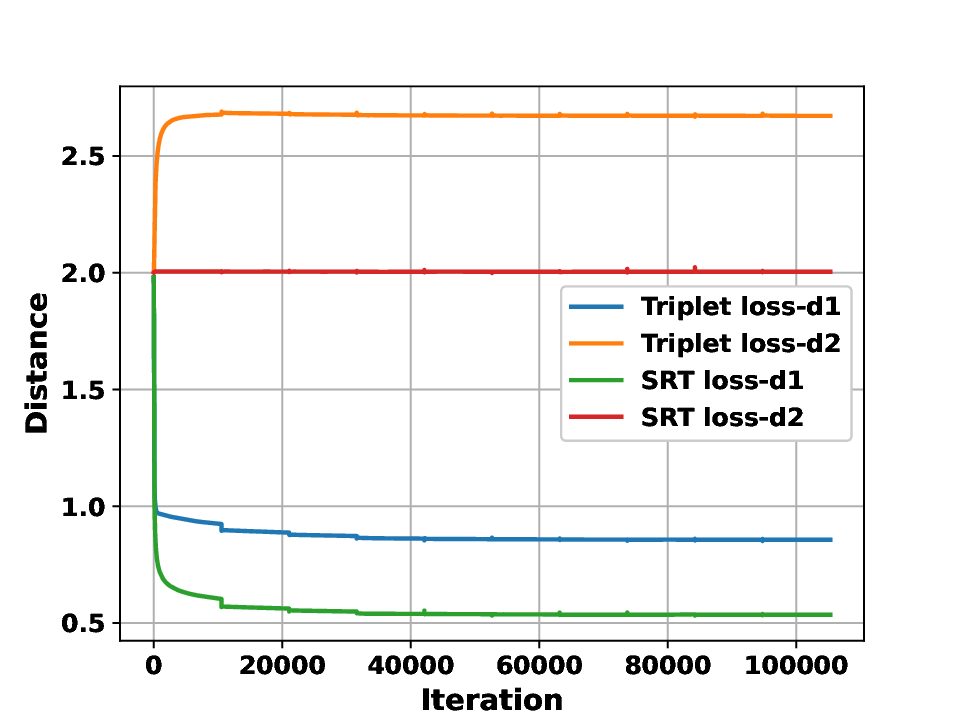}
         \caption{ResNet-50}
         \label{fig:triple_vs_srt_resnet50}
     \end{subfigure}
     \begin{subfigure}[b]{0.30\textwidth}
         \centering
         \includegraphics[width=\textwidth]{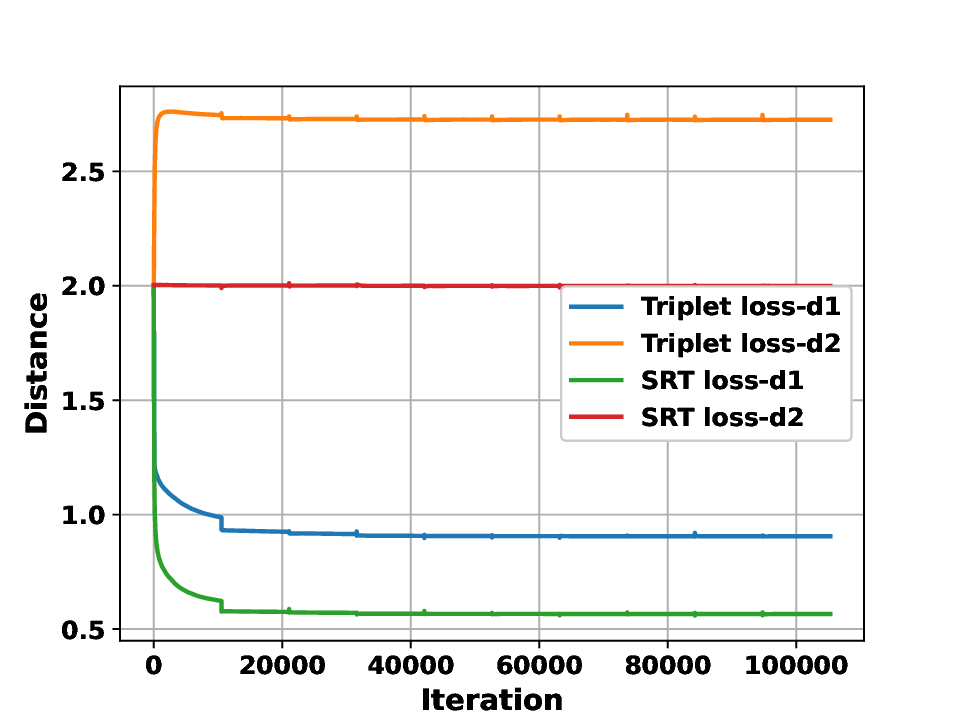}
         \caption{MobileFaceNet}
         \label{fig:triple_vs_srt_mobilefacenet}
     \end{subfigure}
        \caption{
        Naive triplet loss vs. \gls{srt} loss distance learning over training iterations. The plots show the learned d1 (distance between genuine pairs) and d2 (distance between imposter pairs) by each loss over training iterations. It can be clearly noticed that the anchor (model output) of the model trained with \gls{srt} loss is more similar to the positive than the anchor of the model trained with naive triple loss.        }
        \label{fig:triple_vs_srt}
\end{figure*}
\subsubsection{Self-restrained Triplet Loss}
\label{sec:srt}
\begin{figure*}[ht!]
    \centering
    \includegraphics[width=0.77\textwidth]{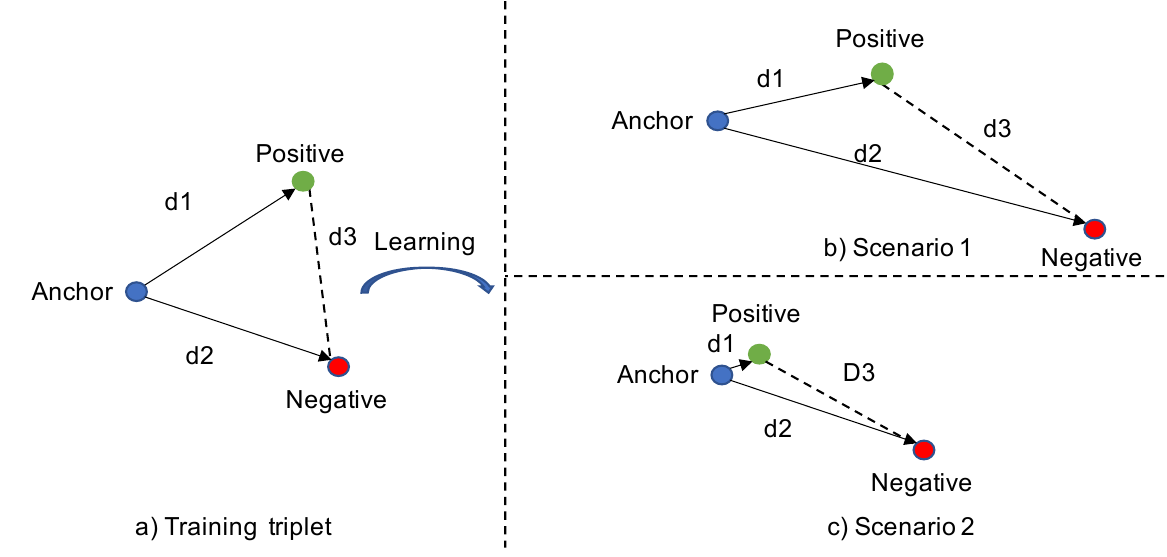}
    \caption{Triplet loss guides the model to maximize the distance between the anchor and negative such as it is greater than the distance between the anchor and positive with the addition of a fixed margin value. One can be clearly noticed the high similarity between the anchor and positive (d1) learned in scenario 2, in comparison to the d1 learned one in scenario 1, whereas, the distance, d2, between the anchor and the negative (imposter pairs) in scenario 1 is extremely large than the d2 in scenario 2.  }
    \label{fig:triplet}
\end{figure*}

Previous works \cite{DBLP:journals/corr/SchroffKP15,DBLP:conf/icip/FengWH00W20} indicated that utilizing triplet-based learning is beneficial for learning discriminative face embeddings.
Let $x \in X$ represents a batch of training samples, and $f(x)$ is the face embeddings obtained from the face recognition model. 
Training with triplet loss requires a triplet of samples in the form $\{x^a_{i},x^p_{i},x^n_{i}\} \in X$, where $x^a_{i}$, the anchor, and $x^p_{i}$, the positive, are two different samples of the same identity, and $x^n_{i}$, the negative, is a sample belonging to a different identity.
The learning objective of the triplet loss is that the distance between $f(x^a_{i})$ and $f(x^p_{i})$ (genuine pairs) with the addition of a fixed margin value (m) is smaller than the distance between $f(x^a_{i})$ and any face embedding $f(x^p_{i})$ of any other identities (imposter pairs). In FaceNet \cite{DBLP:journals/corr/SchroffKP15}, triplet loss is proposed to learn face embeddings by applying the Euclidean distance to normalized face embeddings. Formally, the triplet loss $\ell_t$ for a  mini-batch of $N$ samples is defined as follow:
\begin{equation}
\label{eq:triplet}
    \ell_t= \frac{1}{N} \sum_i^N \max \{d(f(x^a_{i}), f(x^p_{i})) - d(f(x^a_{i}), f(x^n_{i})) + {\rm m}, 0\},
\end{equation}
where $m$ is a margin applied to impose the separability between genuine and imposter pairs. An $d$ is the euclidean distance applied on normalized features and it is given by:
\begin{equation}
\label{eq:l2}
     d(x_i,y_i)=\left\lVert {\bf x}_i - {\bf y}_i \right\rVert_2^2.
\end{equation}
Figure \ref{fig:triplet} visualize two triplet loss learning scenarios.  Figure \ref{fig:triplet}.a shows the initial training triplet, and Figure \ref{fig:triplet}.b and \ref{fig:triplet}.c illustrate two scenarios that can be learnt using triplet loss. In both scenarios, the goal of the triplet loss is achieved i.e. $d(f(x^a_{i}),f(x^n_{i}))> d(f(x^a_{i}),f(x^p_{i}))+ m$. 
In Figure \ref{fig:triplet}.b (scenario 1), both distances are optimized. However,
in this scenario, the optimization of d2 distance is greater than the optimization of d1 distance. 
Whereas, in  Figure \ref{fig:triplet}.c (scenario 2), the triplet loss enforces the model to focus on minimizing the distance between the anchor and the positive.  The optimal state for the triplet loss is achieved when both distance are fully optimized i.e. $d(f(x^a_{i}),f(x^p_{i}))$ is equal to zero and $d(f(x^a_{i}),f(x^n_{i}))$ is greater than the predefined margin. However, achieving such a state may not be feasible, and it requires a huge number of training triplets with large computational resources for selecting the optimal triplets for training.
Given a masked face embedding,  our model is trained to generate a new embedding such as it is similar to the unmasked face embedding of the same identity and dissimilar from other face embeddings of any other identities. As discussed earlier in this section, the distance between imposter pairs is less affected by wearing a mask \cite{DBLP:conf/biosig/DamerGCBKK20,ngan2020ongoing}. Thus, we aim to ensure that our proposed loss focuses on minimizing the distance between the genuine pairs (similar to scenario 2) while maintaining the distance between imposter pairs.

Training \gls{eum} with \gls{srt} loss requires a triple to be defined as follows: $f(x^a_{i})$ is an anchor of masked face embedding, $EUM(f(x^a_{i}))$ is the anchor given as an output of the EUM, $f(x^p_{i})$ is a positive of unmasked embedding, and $f(x^n_{i})$ is a negative embedding of a different identity than anchor and positive. This triplet is illustrated in Figure \ref{fig:workflow}. We want to ensure  that  the distance (d1) between $EUM(f(x^a_{i}))$ and $f(x^p_{i})$ in addition to a predefined margin is smaller than the distance (d2) between $EUM(f(x^a_{i}))$ and $f(x^n_{i})$. 
Our goal is to train EUM to focus on minimizing d1, as d2 is less affected by the mask. 

Under the assumption that the distance between the positive and the negative embeddings (d3) is close to optimal and it does not contribute to the back-propagation of the EUM model, we propose to use this distance as a reference to control the triplet loss. The main idea is to train the model as a naive triplet loss when d2 (anchor-negative distance) is smaller than d3 (positive-negative distance). In this case, the \gls{srt} guides the model to maximize d2 distance and to minimize d1 distance. When d2 is equal or greater than d3, we replace d2 with d3 in the loss calculation. This distance swapping allows the \gls{srt} to learn only, at this point, to minimize d1 distance.
At any point of the training, when the condition on d2 is violated i.e $d(d2)<d(d3)$, the \gls{srt} behave again as a naive triplet loss. We opt to compare the d2 and d3 distances on the batch level to avoid swapping the distance on every minor update on the distance between the imposter pairs.
In this case, we want to ensure that the d1 distance, with the addition of a margin $m$, is smaller than the mean of the d3 distances calculated on the mini-batch of triplets.
Thus, our loss is less sensitive to the outliers resulting from comparing imposter pairs. 
We define our \gls{srt} loss for a mini-batch of the size $N$ as follow:
\begin{equation}
\label{equ:srt}
\resizebox{.9\hsize}{!}{$
   \ell_{SRT}=
   \begin{cases}
   \frac{1}{N} \sum_i^N \max \{d(f(x^a_{i}), f(x^p_{i})) - d(f(x^a_{i}), f(x^n_{i})) + {\rm m}, 0\} & \text{if }  \mu(d2) < \mu(d3)\\
   \frac{1}{N} \sum_i^N \max \{d(f(x^a_{i}), f(x^p_{i})) - \mu(d3) + {\rm m}, 0\} &\text{otherwise} ,\\
   \end{cases}$}
\end{equation}
where $\mu(d2)$ is the mean of the distances between the anchor and the negative pairs calculated on the mini-batch level, given as
$\frac{1}{N} \sum_i^N(d(f(x^a_{i}),f(x^n_{i}))$.
$\mu(d3)$ is the mean of the distances between the positive and the negative pairs calculated on the mini-batch level, given as
$\frac{1}{N} \sum_i^N(d(f(x^p_{i}),f(x^n_{i}))$. An $d$ is the euclidean distance computed on normalized feature embedding (Equation \ref{eq:l2}).

Figure \ref{fig:triple_vs_srt} illustrates the optimization of d1 (distance between genuine pairs) and d2 (distance between imposter pairs) by naive triplet loss and \gls{srt} loss over the training iterations of three EUM models on top of ResNet-100 (Figure \ref{fig:triple_vs_srt_resnet100}), ResNet-50 (Figure \ref{fig:triple_vs_srt_resnet50}) and MobileFaceNet (Figure \ref{fig:triple_vs_srt_mobilefacenet}).
Details on the training settings are presented in Section \ref{sec:exp}. 
It can be clearly noticed that the d1 distance (anchor-positive distance) learned by \gls{srt} is significantly smaller than the one learned by naive triplet loss. This indicates that the output embedding of the \gls{eum} trained with \gls{srt} is more similar to the embedding of the same identity (the positive) than the output embedding of \gls{eum} trained with triplet loss.




\section{Experimental setup}
\label{sec:exp}
This section presents the experimental setups and the implementation details applied in the paper. 
\subsection{Face Recognition Model}
To provide a deep evaluation of the performance of the proposed solution, we evaluated our proposed solution on top of three face recognition models - ResNet-100 \cite{DBLP:conf/cvpr/HeZRS16},  ResNet-50 \cite{DBLP:conf/cvpr/HeZRS16} and MobileFaceNet \cite{DBLP:journals/corr/abs-1804-07573}.
ResNet is one of the widely used Convolutional Neural Network (CNN) architecture used by several face recognition models, e.g. ArcFace \cite{DBLP:conf/cvpr/DengGXZ19} and VGGFace2 \cite{DBLP:conf/fgr/CaoSXPZ18}. 

MobileFaceNet is a compact model designed for low computational powered devices. MobileFaceNet model architecture is based on residual  bottlenecks proposed by MobileNetV2 \cite{DBLP:journals/corr/abs-1801-04381} and depth-wise separable convolutions layer, which allows building a CNN model with a much smaller set of parameters in comparison to standard CNNs. To provide fair and comparable evaluation results, ResNet-50 and MobileFaceNet are trained using the same loss function, the Arcface loss \cite{DBLP:conf/cvpr/DengGXZ19}, and the same training dataset, MS1MV2 \cite{DBLP:conf/cvpr/DengGXZ19}. 
The MS1MV2 is a refined version of the MS-Celeb-1M \cite{DBLP:conf/eccv/GuoZHHG16} dataset.
For ResNet-100, we use the pretrained model released by \cite{DBLP:conf/cvpr/DengGXZ19}. ResNet-100 is trained with ArcFace loss on MS1MV2 \cite{DBLP:conf/cvpr/DengGXZ19}.
Our choice is to employ Arcface loss as it achieved state-of-the-art performance of several face recognition testing datasets such as Labeled Face in the Wild (LFW) \cite{LFWTech}.
The achieved accuracy on LFW by ResNet-100, ResNet-50  and MobileFaceNet trained with Arcface loss using MS1MV2 dataset are  99.83\%,  99.80\%, and 99.55\%, respectively.
The considered face recognition models are evaluated with cosine-distance for comparison. 
The Multi-task Cascaded Convolutional Networks (MTCNN) solution \cite{zhang2016joint} is employed to detect and align the input face image.  All models process aligned and cropped face image of size $112 \times 112$ pixels to produce $512-D$ embedding feature by  ResNet-100 and ResNet-50 and $128-D$ embedding feature by MobileFaceNet.

\begin{figure*}[ht!]
     \centering
     \begin{subfigure}[b]{0.29\textwidth}
         \centering
         \includegraphics[width=\textwidth]{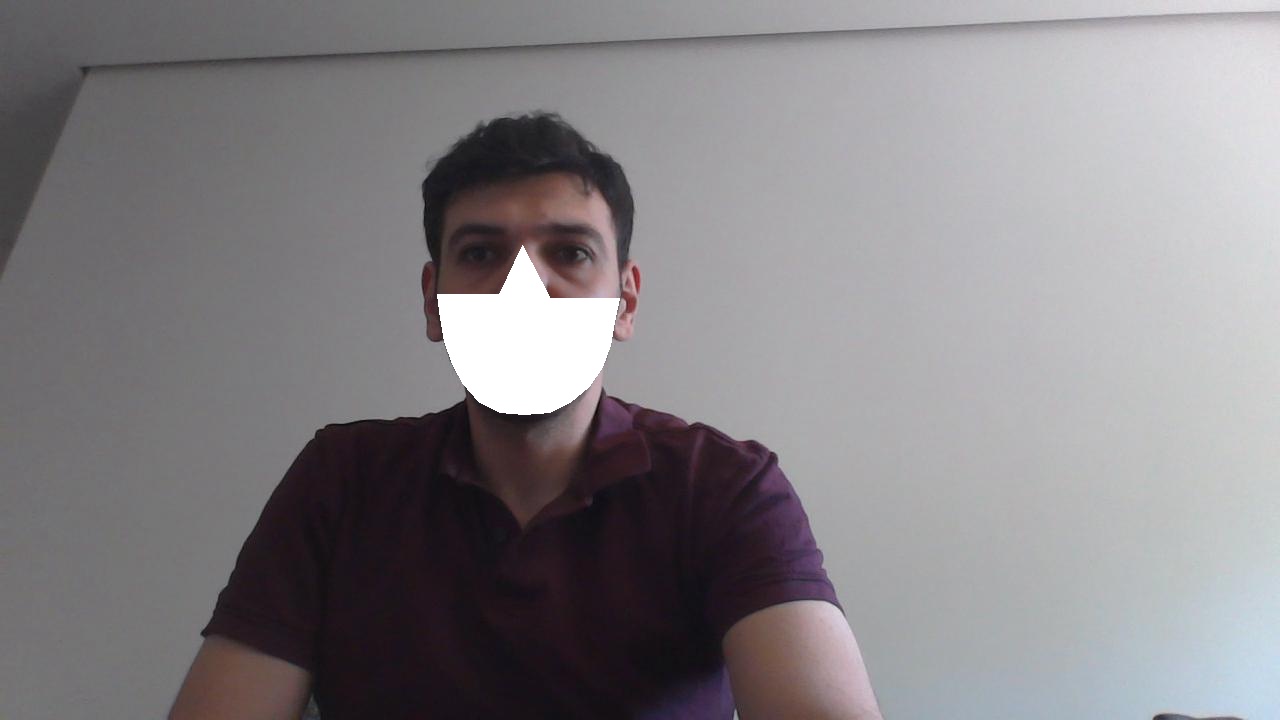}
         \caption{Wide-high coverage}
         \label{fig:mask_a}
     \end{subfigure}
       \begin{subfigure}[b]{0.29\textwidth}
         \centering
         \includegraphics[width=\textwidth]{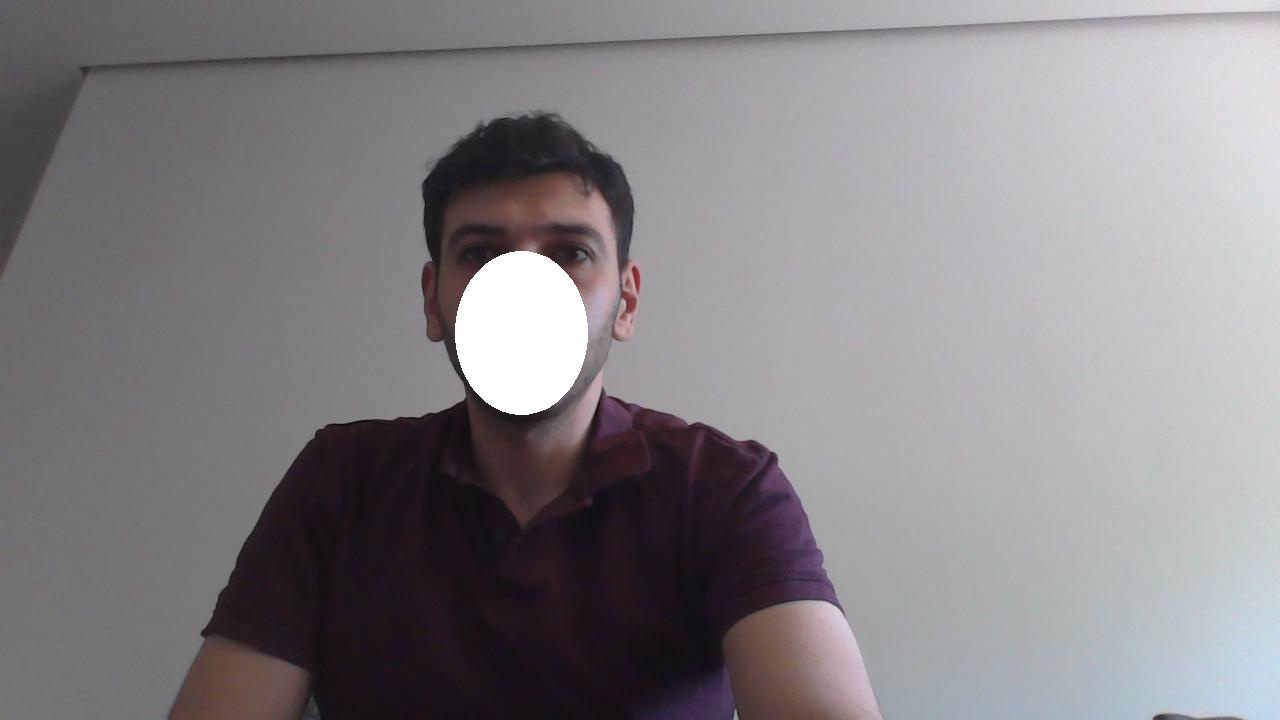}
         \caption{Round-high coverage}
         \label{fig:mask_b}
     \end{subfigure}
          \begin{subfigure}[b]{0.29\textwidth}
         \centering
         \includegraphics[width=\textwidth]{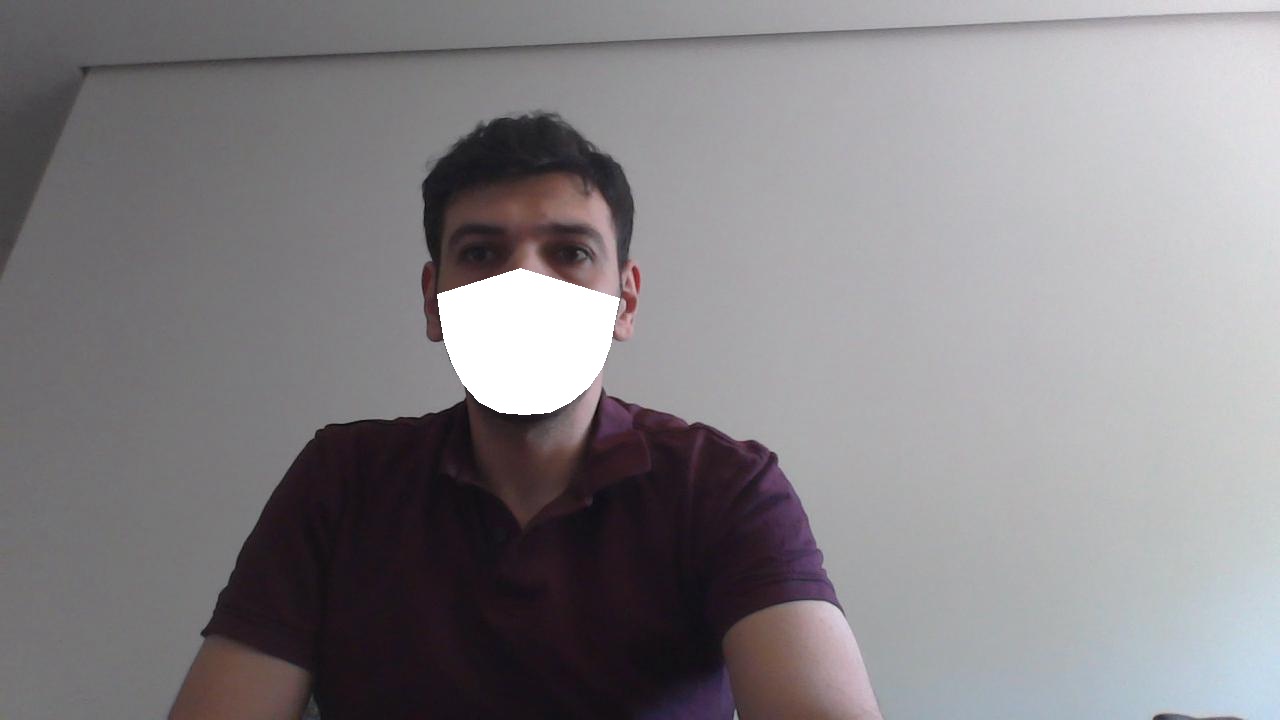}
         \caption{Wide-medium coverage}
         \label{fig:mask_c}
     \end{subfigure}
     
          \begin{subfigure}[b]{0.29\textwidth}
         \centering
         \includegraphics[width=\textwidth]{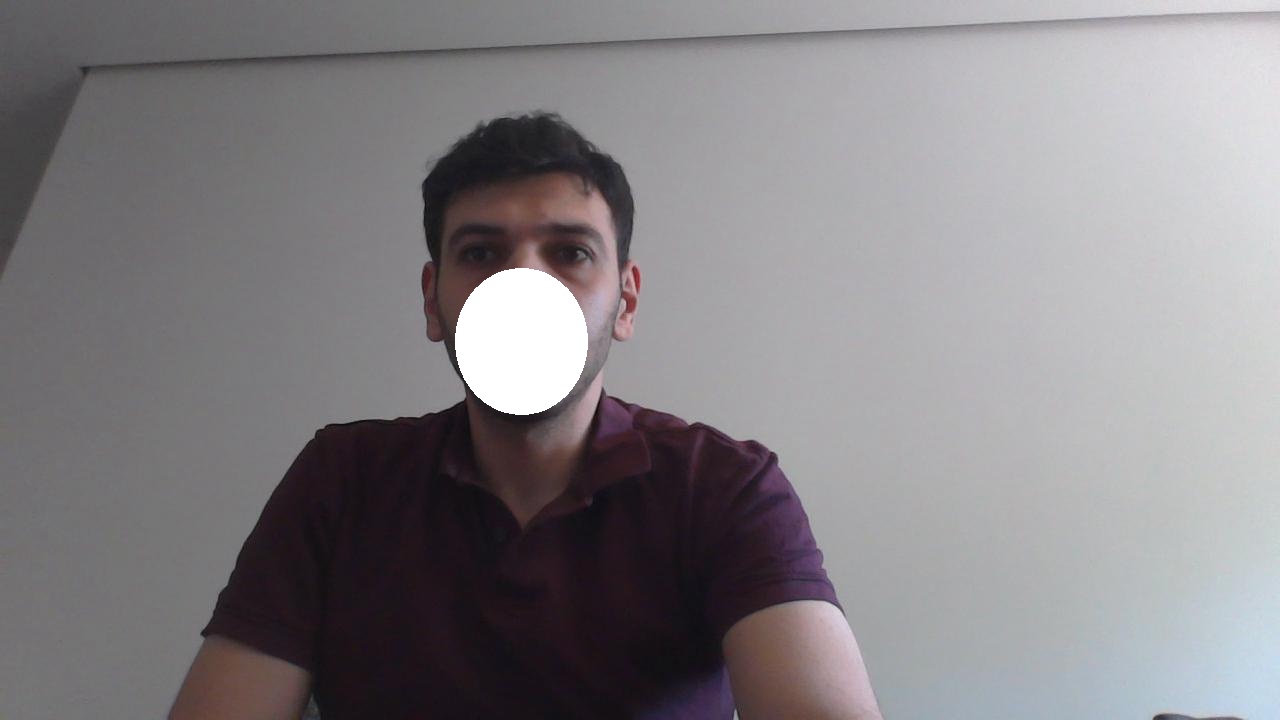}
         \caption{Round-medium coverage}
         \label{fig:mask_d}
     \end{subfigure}
          \begin{subfigure}[b]{0.29\textwidth}
         \centering
         \includegraphics[width=\textwidth]{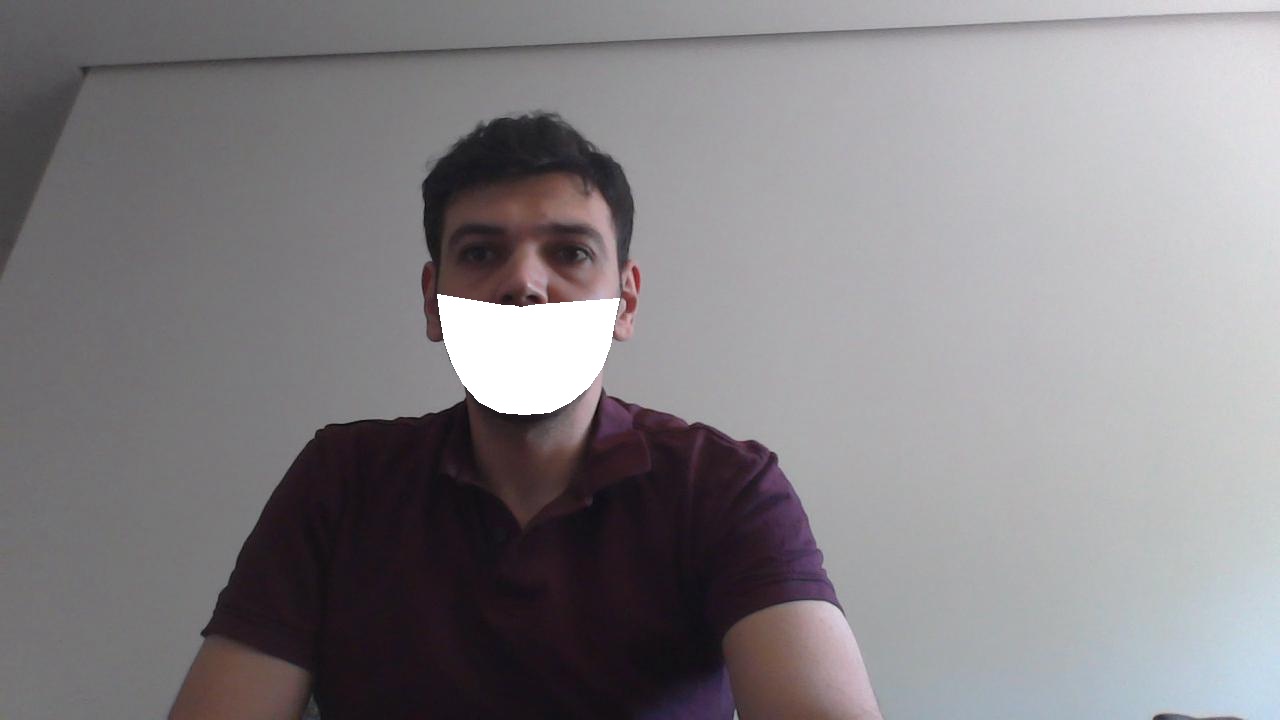}
         \caption{Wide-low coverage}
         \label{fig:mask_e}
     \end{subfigure}
          \begin{subfigure}[b]{0.29\textwidth}
         \centering
         \includegraphics[width=\textwidth]{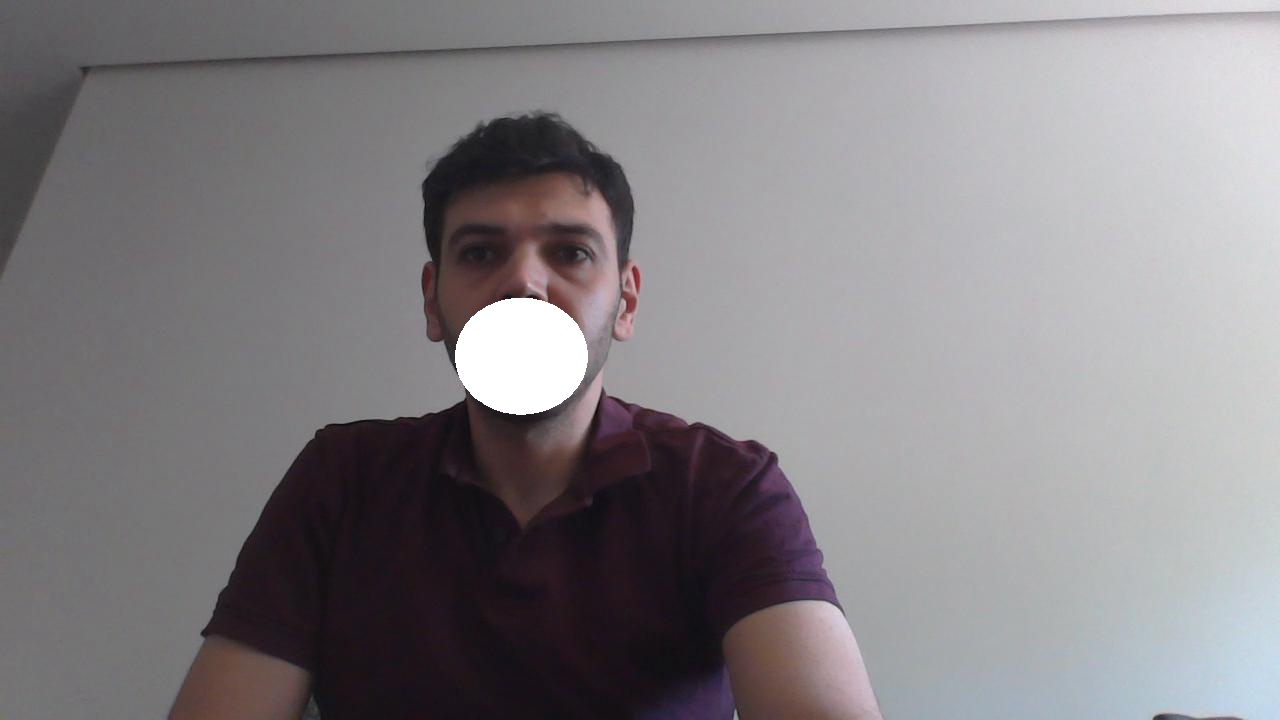}
         \caption{Round-low coverage}
         \label{fig:mask_f}
     \end{subfigure}
        \caption{Samples of the synthetically generated face masks of different shape and coverage. }
        \label{fig:simulated_mask}
\end{figure*}

\subsection{Synthetic Mask Generation}
\label{sec:syn}
As there is no large-scale dataset with pairs of unmasked and masked face images, we opted to use a synthetically generated mask to train our proposed approach.
Specifically, we use the synthetic mask generation method proposed by NIST \cite{ngan2020ongoing}. The synthetic mask generation method depends on the Dlib \cite{DBLP:journals/jmlr/King09} Toolkit for detecting and extracting $68$ facial landmarks from a face image. Based on the extracted landmark points, a face mask of different shapes, heights and colors can be drawn on the face images. The detailed implementation of the synthetic mask generation method is described in \cite{ngan2020ongoing}. The synthetic mask generation method provided six mask types with different heights and coverage: A) wide-high coverage, B) round-high coverage, C) wide-medium coverage, D) round-medium coverage, E) wide-low coverage, and F) round-low coverage. Figure \ref{fig:simulated_mask} shows examples of the simulated face mask with different types and coverage levels.
To synthetically generate a masked face image, we first extract the facial landmark points of the input face image. Then, a mask with a specific color and type can be drawn on the face image using the $x,y$ coordinates of the facial landmarks points. 

\subsection{Dataset}
\label{sec:db}
We used MS1MV2 dataset \cite{DBLP:conf/cvpr/DengGXZ19} to train our proposed approach. The MS1MV2 is a refined version of MS-Celeb-1M \cite{DBLP:conf/eccv/GuoZHHG16} dataset. The MS1MV2 contains $58m$ images of $85k$ different identities. We generated a masked version of the MS1MV2 noted as MS1MV2-Masked as described in Section \ref{sec:syn}. The mask type (as described in Section \ref{sec:syn}) and color are randomly selected for each image to add more diversity of mask color and coverage to the training dataset. The Dlib failed in extracting the facial landmarks from $426k$ images. These images are neglected from the training dataset.
A subset of $5k$ images are randomly selected from  MS1MV2-Masked to validate the model during the training phase. 

The proposed solution is evaluated using two real masked face datasets: Masked Faces in Real World for Face Recognition (MRF2) \cite{anwar2020masked} and Masked face recognition (MFR) \cite{DBLP:conf/biosig/DamerGCBKK20,IET_B_Mask}.
We also evaluated our solution on two larger-scale datasets with synthetically generated masks. We use the synthetic mask generation method described in Section \ref{sec:syn} (proposed by NIST \cite{ngan2020ongoing}) to synthetically generate masked faces from the Labeled Faces in the Wild (LFW) \cite{LFWTech} and IARPA Janus Benchmark -C (IJB-C) \cite{DBLP:conf/icb/MazeADKMO0NACG18}. The mask type and color are randomly selected for each image in the LFW and IJB-C datasets to achieve a greater variety of mask types and colors. In the following, we briefly describe each of the evaluation datasets used in this work.

\paragraph{Masked Faces in Real World for Face Recognition (MRF2)}  MFR2 \cite{anwar2020masked} contains 269 images of 53 identities crawled from the internet. Therefore, the images of the MRF2 dataset can be considered as captured under in-the-wild conditions.
The dataset contains images of real masked and unmasked faces with an average of 5 images per identity.
\paragraph{Masked Face Recognition (MFR)} We deploy an extended version of the MFR dataset \cite{DBLP:conf/biosig/DamerGCBKK20,IET_B_Mask}. The extended version of MFR is collected from 48 participants using their webcams under three different sessions- session 1 (reference) and session 2 and 3 (probes). The sessions are captured on three different days.  Each session contains data captured using three videos. In each session, the first video is recorded when the subject is not wearing a facial mask in the daylight without additional electric lighting. The second and third videos are recorded when the subject is wearing a facial mask and with no additional electric lighting in the second video and with electric lighting in the third video (room light is turned on). The first session (reference) contains 480 unmasked images and 960 masked images. The second and the third sessions (probe) contain 960 unmasked images and 1920 masked images.
\paragraph{Labeled Faces in the  Wild (LFW)}
LFW \cite{LFWTech} is an unconstrained face verification benchmark. It contains 13,233 images of 5,749 identities. The number of comparison pairs in unrestricted with labeled outside data protocol \cite{LFWTech} of LFW is 6000 (3000 genuine and 3000 imposter comparisons).
\paragraph{IARPA Janus Benchmark–C (IJB-C)}
IJB-C dataset \cite{DBLP:conf/icb/MazeADKMO0NACG18} is one of the largest face verification benchmark. IJB-C consists of 31,334 still images and 117,542 frames of 11,779 videos of 3531 identities. The 1:1 mixed verification protocol \cite{DBLP:conf/icb/MazeADKMO0NACG18} of IJB-C contains 19,557 genuine and 15,638,932 impostor comparisons. 
\subsection{Evaluation Settings}
We reported the verification performances for each of the evaluation datasets under seven experimental settings. Also, for each of the conducted experiments, we report
the failure to extract rate (FTX) to capture the effect of wearing a face mask on face detection.
FTX measure is the proportion of comparisons where the feature extraction was not possible. For IJB-C and LFW, we used the bounding box provided by the datasets to align and crop the face. Therefore, the FTX for LFW and IJB-C are 0.0\% in all experimental settings.
For MFR and MRF2 datasets, we reported the FTX for each of the experiment settings.
The conducted experiments are defined as follow:
\paragraph{Unmasked Reference-Unmasked Probe (UMR-UMP)} The unmasked references are compared to unmasked probes.
For LFW and IJB-C, we followed the evaluation protocol given by each of these datasets and evaluated them based on the provided comparison pairs.  The number of genuine comparisons is 3000 in LFW and 19,557 in IJB-C.  The number of imposter comparisons is 3000 in LFW and 15,638,932 in IJB-C.
The evaluation of UMR-UMP on the MFR2 dataset is done by performing N:N comparisons between all unmasked faces resulting in 90 genuine and 9,416 imposter comparisons. 
For the MFR dataset, we performed N:M comparisons between the unmasked reference of the first session (reference session) and unmasked probe of the second and the third sessions (probe sessions) resulting in 9,600 genuine and 451,200 imposter comparisons. The FTXs of MFR and MRF2 when the probes and the references are unmasked are 0.0\%.
\paragraph{Unmasked Reference-Masked Probe (UMR-MP)}  The unmasked references, in this case, are compared to masked probes. For LFW and IJB-C datasets, we utilized the exact comparison pairs defined in UMR-UMP experimental settings. Different from UMR-UMP, the probes, in this case, are synthetically masked (as described in section \ref{sec:syn}). We considered the first image in defined pairs as a reference and the second image is considered as a probe. For the MRF2 dataset, we performed N:M comparisons between unmasked and masked sets resulting in 296 genuine and 15090 imposter comparisons.  The FTX of MRF2, in this setting, is 0.9497\%. For the MFR dataset, we performed N:M comparisons between unmasked references of the first session and masked probes of the second and the third sessions. The FXT, in this case, is 4.4237\%, and the number of comparisons is 16,490 genuine and 86,4341 imposter comparisons.
\paragraph{Unmasked Reference-Masked Probe (UMR-MP(T))} The unmasked references are compared to masked probes. Different from UMR-MP, the masked probes are processed by \gls{eum} model trained with conventional triplet loss (T).
\paragraph{Unmasked Reference-Masked Probe (UMR-MP(SRT))} The unmasked references are compared to masked probes processed by \gls{eum} model trained with \gls{srt} loss. In UMR-MP(T) and UMR-MP(SRT), the number of genuine and impostor pairs and the FTXs are identical to UMR-MP experimental setting.
\paragraph{Masked Reference-Masked Probe (MR-MP)} The masked references are compared to masked probes. For LFW and IJB-C, we utilized the same comparison pairs described in UMR-UMP experimental setting. Both reference and probe are synthetically masked. The number of genuine and imposter comparisons, in this case, is the same as in UMR-UMP experimental setting.
For the MRF2 dataset, we performed N:N comparisons between masked faces resulting in 639 genuine and 24,010 imposter comparisons. The FTX, in this case, is  1.2030\% for the MRF2 dataset. For the MFR dataset,  we performed  N:M comparisons between the masked faces of the first session and the masked faces of the second and the third sessions resulting in 31,318 genuine and 1,729,424 imposter comparisons. The FTX, in this case, is 4.4736\% for the MFR dataset.  
\paragraph{Masked Reference-Masked Probe (MR-MP(T))} The masked references are compared to masked probes. Both masked references and probes are processed by \gls{eum} trained with conventional triplet loss (T). The comparison pairs and the FTX are the same as in the MR-MP experimental setting.
\paragraph{Masked Reference-Masked Probe MR-MP(SRT)} The masked references are compared to masked probes. Masked references and probes are processed by \gls{eum} trained with \gls{srt} loss. The comparison pairs and the FTX for all evaluation datasets, in this experimental case,  are identical to the MR-MP case.

\begin{figure*}[ht!]
     \centering
     \begin{subfigure}[b]{0.32\textwidth}
         \centering
         \includegraphics[width=\textwidth]{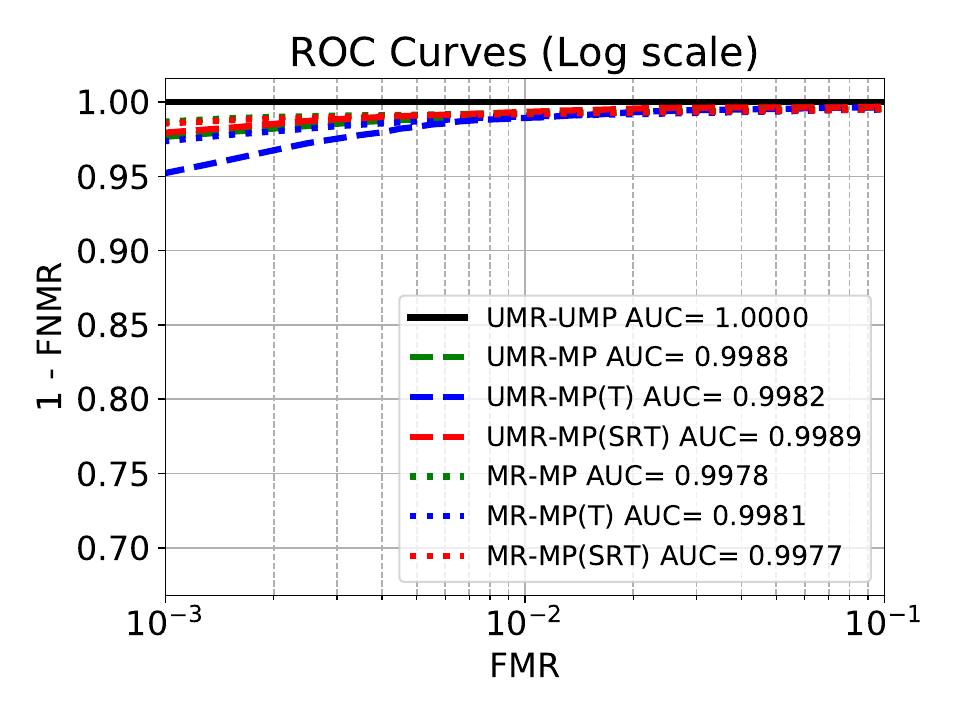}
         \caption{ResNet-100: MFR}
         \label{fig:roc_r100_mfr}
     \end{subfigure}
          \begin{subfigure}[b]{0.32\textwidth}
         \centering
         \includegraphics[width=\textwidth]{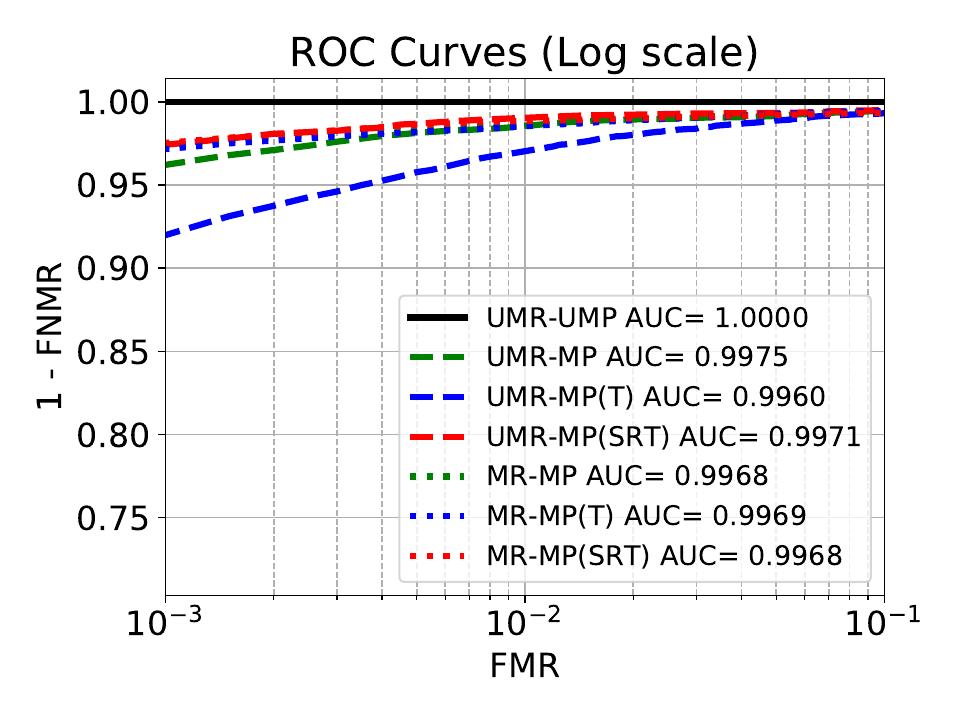}
         \caption{ResNet-50: MFR}
         \label{fig:roc_r50_mfr}
     \end{subfigure}
     \begin{subfigure}[b]{0.32\textwidth}
         \centering
         \includegraphics[width=\textwidth]{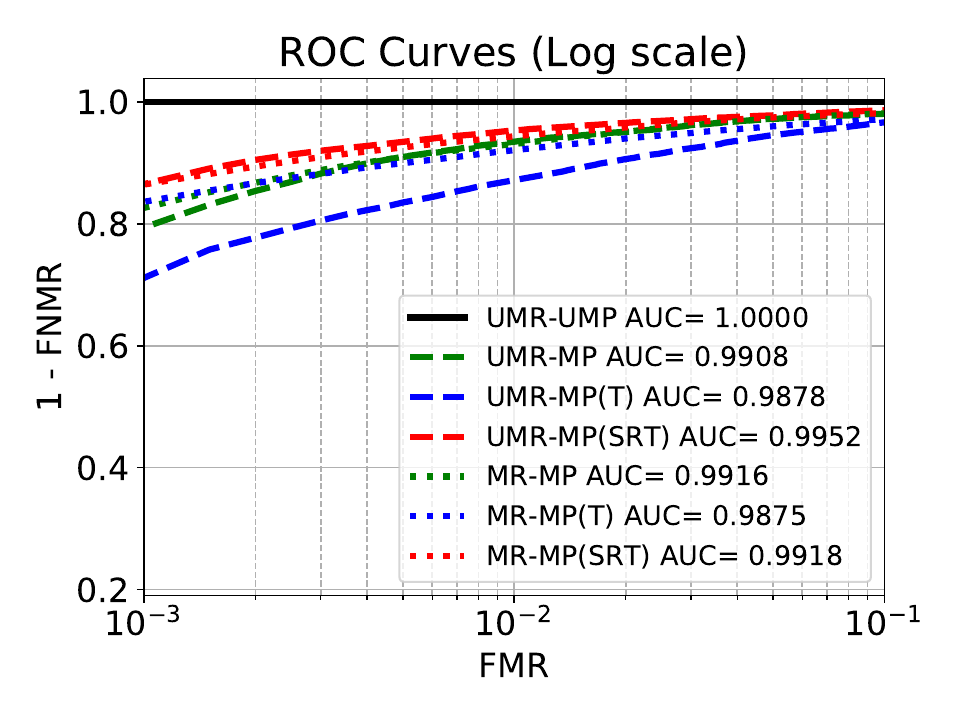}
         \caption{MobileFaceNet: MFR}
         \label{fig:roc_m_mfr}
          \end{subfigure}
    \begin{subfigure}[b]{0.32\textwidth}
         \centering
         \includegraphics[width=\textwidth]{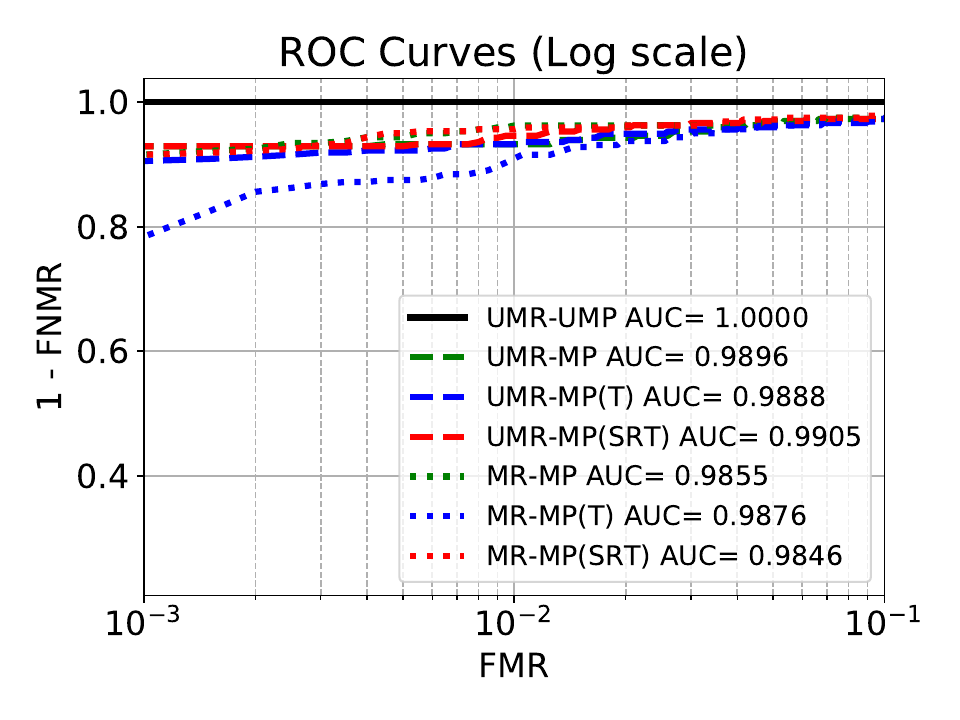}
         \caption{ResNet-100: MRF2}
         \label{fig:roc_r100_mrf2}
     \end{subfigure}
     \begin{subfigure}[b]{0.32\textwidth}
         \centering
         \includegraphics[width=\textwidth]{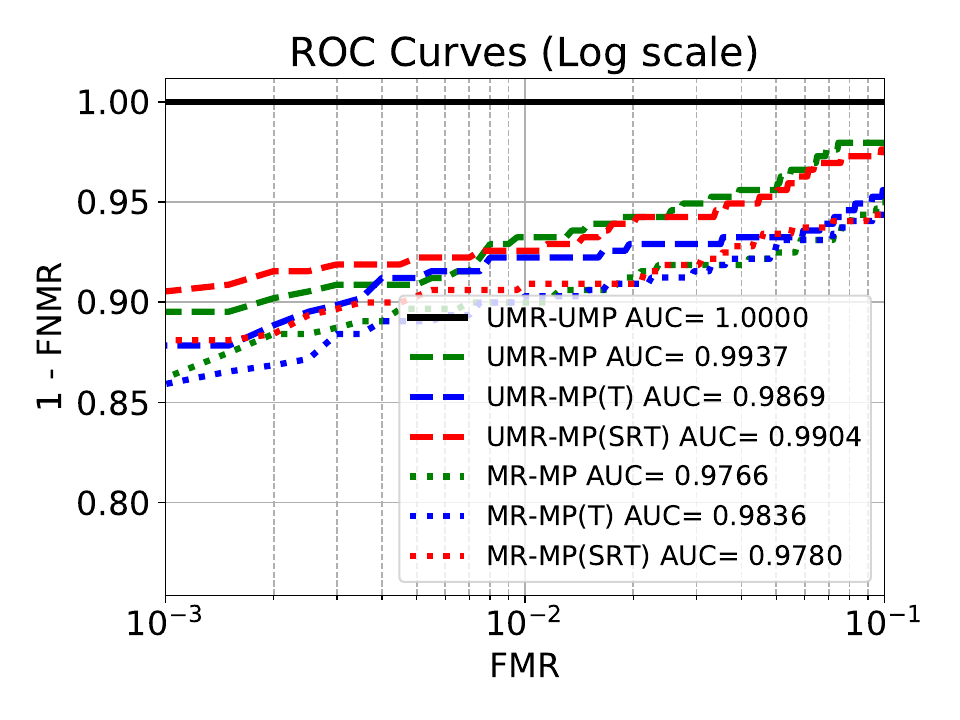}
         \caption{ResNet-50: MRF2}
         \label{fig:roc_r50_mrf2}
     \end{subfigure}
     \begin{subfigure}[b]{0.32\textwidth}
         \centering
         \includegraphics[width=\textwidth]{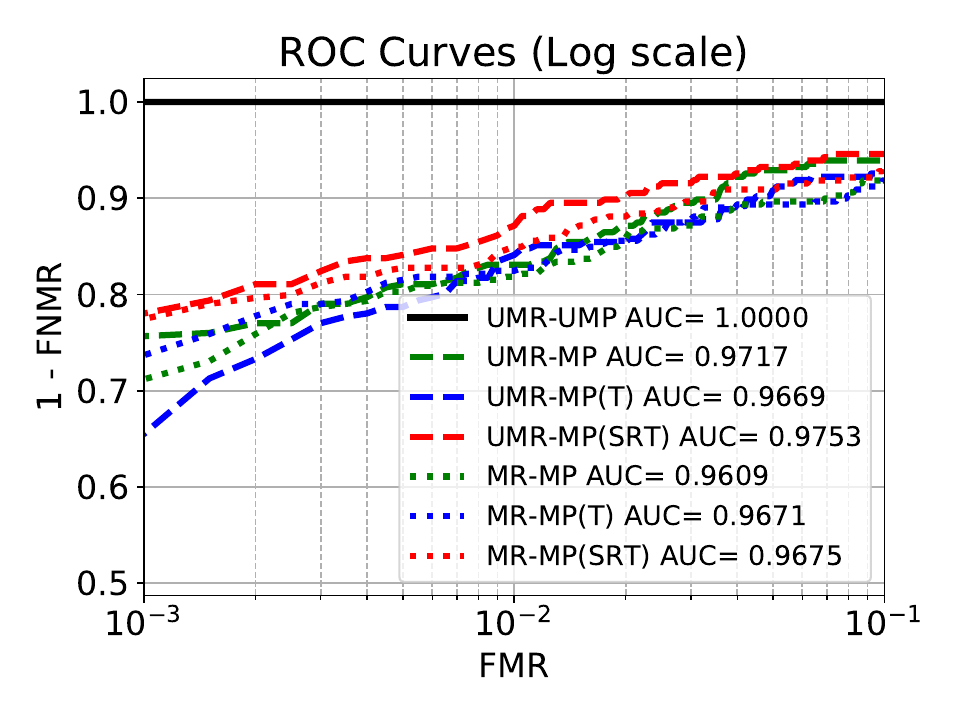}
         \caption{MobileFaceNet: MRF2}
         \label{fig:roc_m_mrf2}
     \end{subfigure}
     
    \begin{subfigure}[b]{0.32\textwidth}
         \centering
         \includegraphics[width=\textwidth]{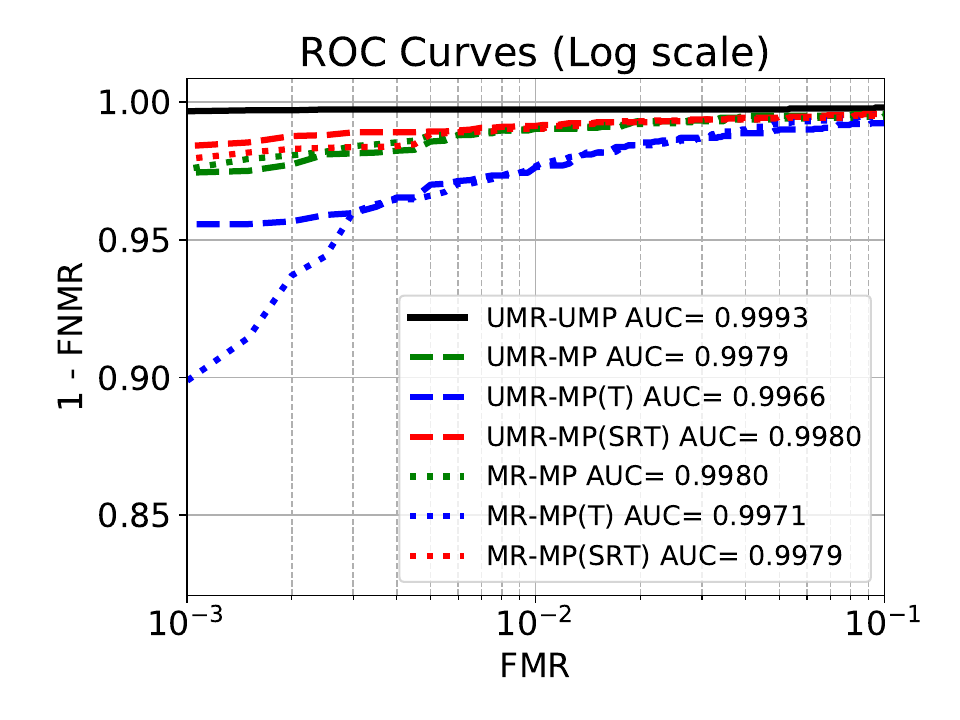}
         \caption{ResNet-100: LFW}
         \label{fig:roc_r100_lfw}
     \end{subfigure}
     \begin{subfigure}[b]{0.32\textwidth}
         \centering
         \includegraphics[width=\textwidth]{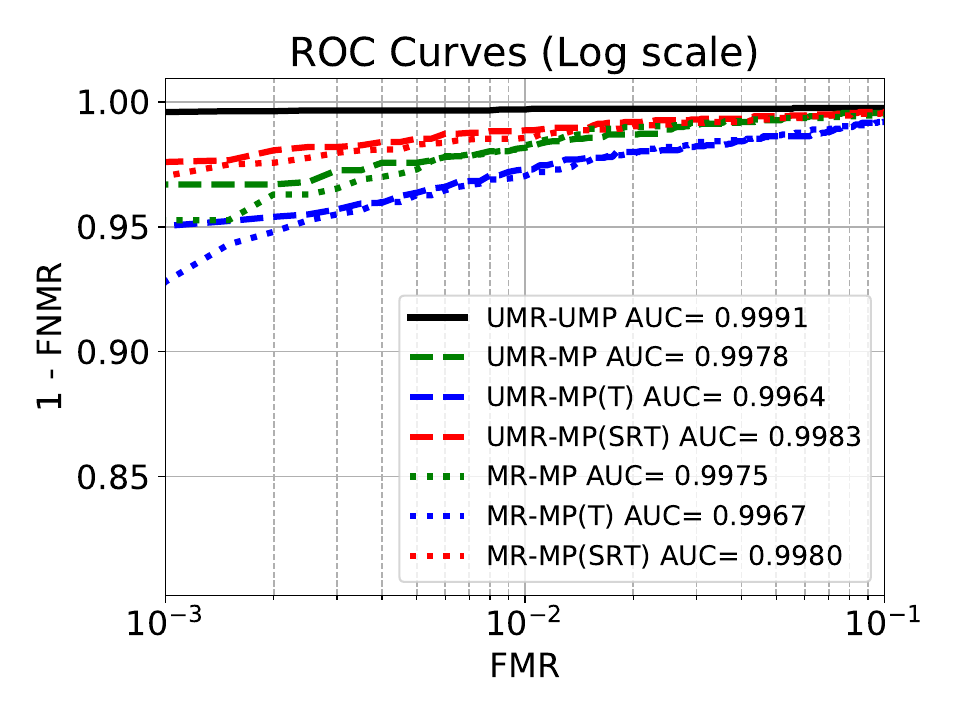}
         \caption{ResNet-50: LFW}
         \label{fig:roc_r50_lfw}
     \end{subfigure}
     \begin{subfigure}[b]{0.32\textwidth}
         \centering
         \includegraphics[width=\textwidth]{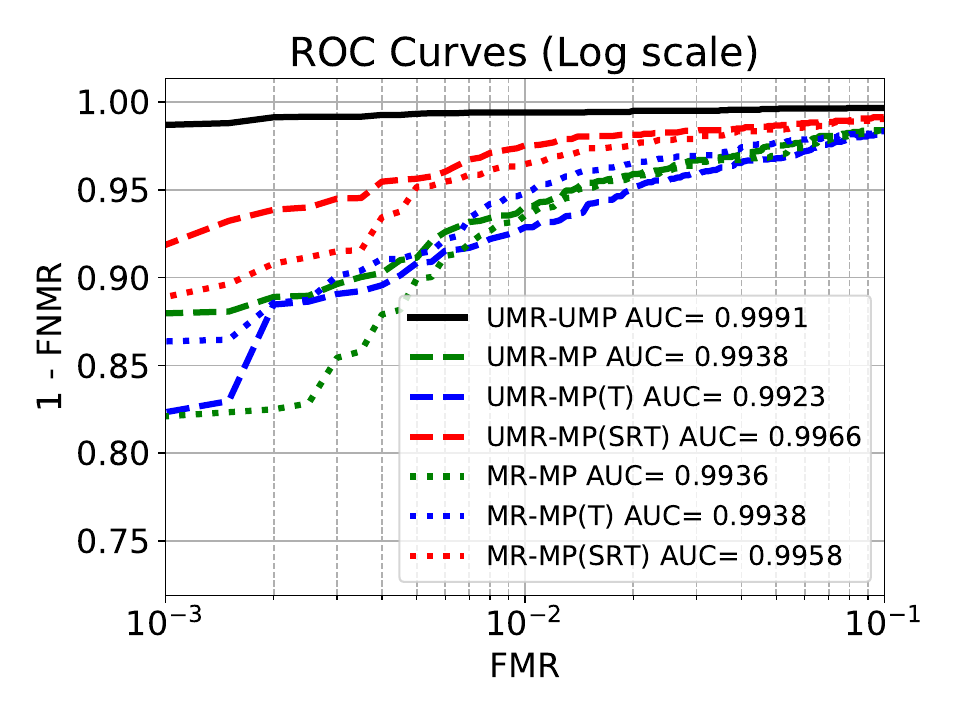}
         \caption{MobileFaceNet: LFW}
         \label{fig:roc_m_lfw}
     \end{subfigure}
     
     \begin{subfigure}[b]{0.32\textwidth}
         \centering
         \includegraphics[width=\textwidth]{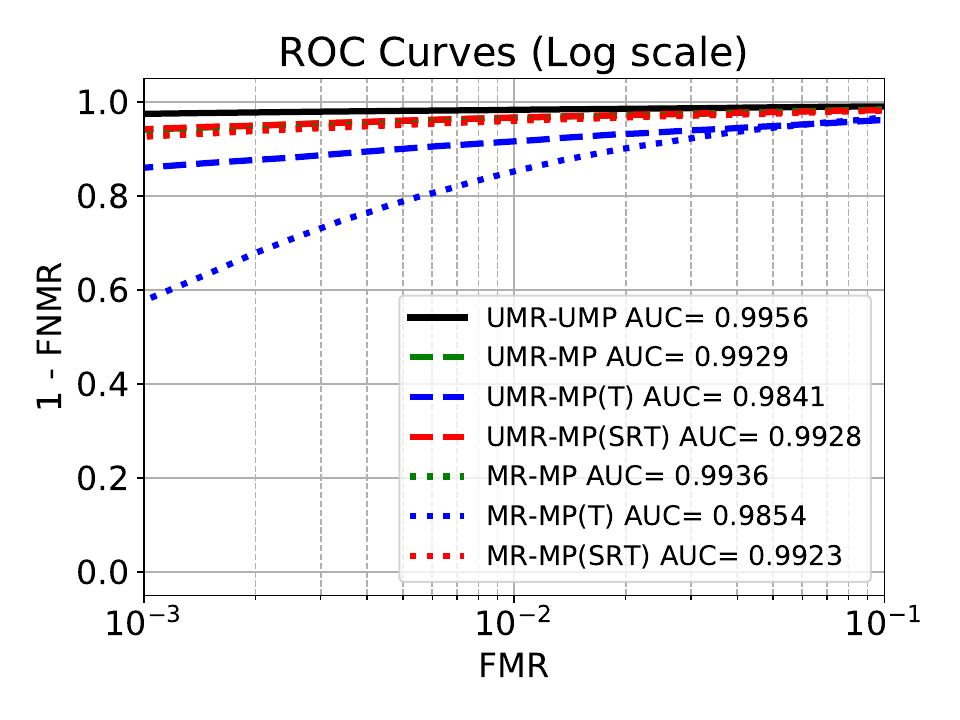}
         \caption{ResNet-100: IJB-C}
         \label{fig:roc_r100_ijbc}
     \end{subfigure}
     \begin{subfigure}[b]{0.32\textwidth}
         \centering
         \includegraphics[width=\textwidth]{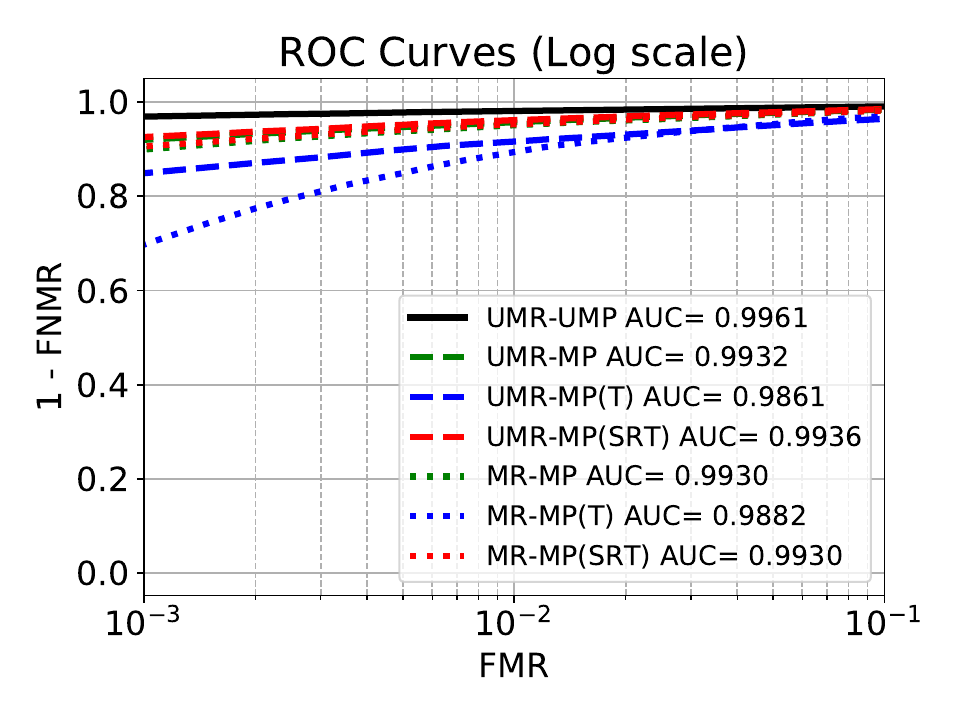}
         \caption{ResNet-50: IJB-C}
         \label{fig:roc_r50_ijbc}
     \end{subfigure}
     \begin{subfigure}[b]{0.32\textwidth}
         \centering
         \includegraphics[width=\textwidth]{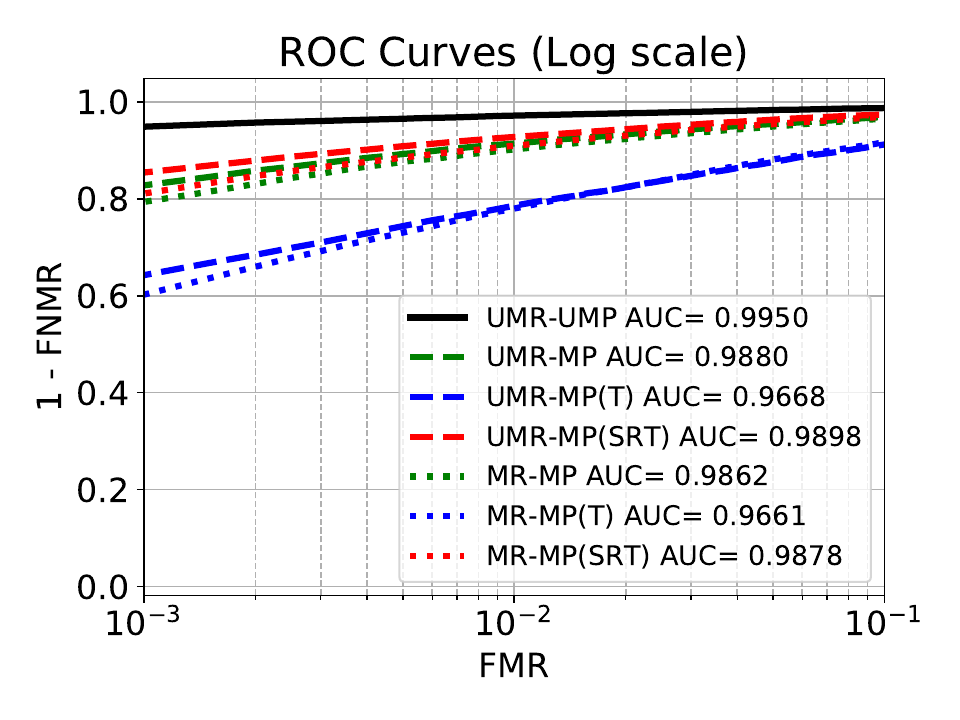}
         \caption{MobileFaceNet: IJB-C}
         \label{fig:roc_m_ijbc}
     \end{subfigure}
        \caption{ 
        The achieved log-scale ROC curves by different experimental settings The ROC curves achieved by \gls{eum} trained with \gls{srt} in all plots are in red. The ROC curves achieved by \gls{eum} trained with the naive triplet in all plots are in blue. The ROC curves of the considered models without \gls{eum} are in green color.  In each plot, the curves of UMR-MP, UMR-MP(T), and UMR-MP(SRT) cases are marked with a dashed line.  The curves of MR-MP, MR-MP(T), and MR-MP(SRT) cases are marked with a dotted line. For each ROC curve, the area under the curve (AUC) is listed inside the plot. 
       }
        \label{fig:roc}
\end{figure*}


\subsection{Model Training Setup}
We trained six instances of the \gls{eum} model. The first, second, and the third instances, ResNet-100 EUM(SRT), ResNet-50 EUM(SRT) and MobileFaceNet EUM(SRT), are trained with \gls{srt} loss using feature embeddings obtained from ResNet-100, ResNet-50 and MobileFaceNet, respectively. The fourth, fifth and sixth instances, ResNet-100 EUM(T), ResNet-50 EUM(T) and MobileFaceNet EUM(T), are trained with triplet loss using feature embeddings obtained from ResNet-100, ResNet-50 and MobileFaceNet, respectively as an ablation study to our proposed \gls{srt}.
The proposed EUM models in this paper are implemented by Pytorch and trained on Nvidia GeForce RTX 2080 GPU. All models are trained using an SGD optimizer with an initial learning rate of 1e-1 and batch size of 512.  The learning rate is divided by 10 at $30k, 60k, 90k$ training iterations. 
The early-stopping patience parameter is set to 3 (around 30k training iteration) causing ResNet-100 EUM(SRT), ResNet-50 EUM(SRT), MobileFaceNet EUM(SRT), ResNet-100 EUM(T), ResNet-50 EUM(T) and MobileFaceNet EUM(T) to stop after 10k, 80k, 70k, 10k, 60k, 10k training iterations, respectively.

\subsection{Evaluation Metric}
The verification performance is reported as Equal Error Rate (EER), as well as, FMR100, and FMR1000, which are the lowest \gls{fnmr} for a \gls{fmr}$\leq$1.0\% and $\leq$0.1\%, respectively. 
Additionally, we calculate and report the operation thresholds at FMR100 (FMR100\_Th) and FMR1000 (FMR1000\_Th) for each of the evaluated models and each of the benchmarks based on UMR-UMP experimental setting (unmasked reference - unmasked probe). Based on FMR100\_Th and FMR1000\_Th thresholds, we report the FMR, the FNMR, and the average of the FMR and FNMR (Avg) at these thresholds for all experimental settings. This aims to estimate a realistic scenario where the operational threshold is decided on the conventional UMR-UMP performance.
We also report the mean of the genuine scores (G-mean) and the mean of imposter scores (I-mean) to analysis the shifts in genuine and imposter scores distributions induced by wearing a face mask and to demonstrate the improvement in the verification performance achieved by our proposed solution. For each of the evaluation settings, we plot the receiver operating characteristic (ROC) curves, showing FMR100 and FMR1000 clearly by providing a log-scale FMR axis.
Further, we enrich our reported evaluation results by reporting the Fisher Discriminant Ratio (FDR) to provide an in-depth analysis of the separability of genuine and imposters scores for different experimental settings. FDR is a class separability criterion described in \cite{poh2004study}, and it is given by:
\begin{equation}
\label{eq:fdr}
    FDR=\frac{(\mu_G - \mu_I)^2}{(\sigma_G)^2+(\sigma_I)^2},
\end{equation}
where $\mu_G$ and $\mu_I$ are the genuine and imposter scores mean values and $\sigma_G$ and $\sigma_I$ are their standard deviations values.  The larger the FDR value, the higher is the separation between the genuine and imposters scores and thus better expected verification performance.

\begin{table*}[ht!]
\centering
\resizebox{\textwidth}{!}{%
\begin{tabular}{|l|l|l|l|l|l|l|l|l|l|l|l|l|l|}
\hline
 &
  \cellcolor[HTML]{FFFFFF}{\color[HTML]{000000} } &
   &
   &
   &
  \multicolumn{3}{l|}{FMR100\_Th\textsuperscript{UMR-UMP}} &
  \multicolumn{3}{l|}{FMR1000\_Th\textsuperscript{UMR-UMP}} &
   &
   &
   \\ \cline{6-11}
\multirow{-2}{*}{\begin{tabular}[c]{@{}l@{} }MFR\end{tabular}} &
  \multirow{-2}{*}{\cellcolor[HTML]{FFFFFF}{\color[HTML]{000000} \begin{tabular}[c]{@{}l@{}}Setting\end{tabular}}} &
  \multirow{-2}{*}{EER\%} &
  \multirow{-2}{*}{FMR100\%} &
  \multirow{-2}{*}{FMR1000\%} &
  FMR\% &
  FNMR\% &
  Avg.\% &
  FMR\% &
  FNMR\% &
  Avg.\% &
  \multirow{-2}{*}{G-mean} &
  \multirow{-2}{*}{I-mean} &
  \multirow{-2}{*}{FDR} \\ \hline
 &
  UMR-UMP &
  0.0000 &
  0.0000 &
  0.0000 &
  1.0000 &
  0.0000 &
  0.5000 &
  0.1000 &
  0.0000 &
  0.0500 &
  0.8534 &
  0.0252 &
  70.7159 \\ \cline{2-14} 
 &
  UMR-MP &
  0.8914 &
  0.8793 &
  2.3347 &
  0.4829 &
  1.1886 &
  0.8358 &
  0.0082 &
  6.0461 &
  3.0272 &
  0.5271 &
  0.0203 &
  15.0316 \\ \cdashline{2-14} 
 &
  UMR-MP(T) &
  1.0430 &
  1.0794 &
  4.7726 &
  0.3084 &
  2.4257 &
  1.3670 &
  0.0000 &
  17.4773 &
  8.7386 &
  0.4331 &
  0.0188 &
  12.0587 \\ \cdashline{2-14} 
 &
  UMR-MP(SRT) &
  \textbf{0.7702} &
  0.6610 &
  2.0558 &
  0.4717 &
  0.9460 &
  \textbf{0.7089} &
  0.0108 &
  4.8029 &
 \textbf{2.4068} &
  0.5379 &
  0.0221 &
  15.9027 \\  \cline{2-14} 
 & 
  MR-MP &
  \textbf{0.8014} &
  0.7695 &
  1.3155 &
  4.3230 &
  0.5971 &
  2.4601 &
  0.4031 &
  0.8685 &
  \textbf{0.6358} &
  0.7314 &
  0.0560 &
  18.7469 \\ \cdashline{2-14} 
 &
  MR-MP(T) &
  0.9598 &
  0.9471 &
  2.6348 &
  16.0855 &
  0.4513 &
  8.2684 &
  2.4656 &
  0.7660 &
  1.6158 &
  0.7415 &
  0.1185 &
  15.2544 \\ \cdashline{2-14} 
\multirow{-7}{*}{\rotatebox[origin=c]{90}{ResNet-100}
} &
  MR-MP(SRT) &
  0.8270 &
  0.8015 &
  1.4433 &
  3.6616 &
  0.6482 &
  \textbf{2.1549} &
  0.3083 &
  0.9994 &
  0.6539 &
  0.7248 &
  0.0486 &
  18.3184 \\ \hline \hline
 &
  UMR-UMP &
  0.0000 &
  0.0000 &
  0.0000 &
  1.0000 &
  0.0000 &
  0.5000 &
  0.1000 &
  0.0000 &
  0.0500 &
  0.8538 &
  0.0349 &
  55.9594 \\  \cline{2-14}
 &
  UMR-MP &
  1.2492 &
  1.4251 &
  3.7780 &
  0.4308 &
  1.9709 &
  1.2008 &
  0.0007 &
  10.6246 &
  5.3126 &
  0.5254 &
  0.0251 &
  12.6189 \\   \cdashline{2-14}   
 &
  UMR-MP(T) &
  1.9789 &
  2.9533 &
  7.9988 &
  0.5626 &
  4.0206 &
  2.2916 &
  0.0000 &
  30.6549 &
  15.3275 &
  0.4401 &
  0.0392 &
  9.4412 \\ \cdashline{2-14} 
 &
  UMR-MP(SRT) &
  \textbf{0.9611} &
  0.9460 &
  2.5652 &
  0.5595 &
  1.2129 &
  \textbf{0.8862} &
  0.0030 &
  7.4591 &
  \textbf{3.7310} &
  0.5447 &
  0.0272 &
  13.4045 \\ \cline{2-14} 
 &
  MR-MP &
  1.2963 &
  1.4145 &
  2.6311 &
  3.7683 &
  0.8302 &
  2.2993 &
  0.2222 &
  2.0467 &
  1.1345 &
  0.7232 &
  0.0675 &
  15.1356 \\ \cdashline{2-14} 
 &
  MR-MP(T) &
  1.3091 &
  1.4560 &
  2.8259 &
  96.3681 &
  0.0000 &
  48.1840 &
  62.1757 &
  0.1980 &
  31.1868 &
  0.8269 &
  0.4169 &
  13.0528 \\ \cdashline{2-14} 
\multirow{-7}{*}{
\rotatebox[origin=c]{90}{ResNet-50}
} &
  MR-MP(SRT) &
  \textbf{1.1207} &
  1.1367 &
  2.4523 &
  3.2837 &
  0.8717 &
  \textbf{2.0777} &
  0.2227 &
  1.8775 &
  \textbf{1.0501} &
  0.7189 &
  0.0557 &
  15.1666 \\ \hline \hline
 &
  UMR-UMP &
  0.0000 &
  0.0000 &
  0.0000 &
  1.0000 &
  0.0000 &
  0.5000 &
  0.1000 &
  0.0000 &
  0.0500 &
  0.8432 &
  0.0488 &
  37.3820 \\ \cline{2-14} 
 &
  UMR-MP &
  3.4939 &
  6.5070 &
  20.5640 &
  0.2723 &
  12.3833 &
  6.3278 &
  0.0088 &
  40.4063 &
  20.2075 &
  0.4680 &
  0.0307 &
  7.1499 \\ \cdashline{2-14} 
 &
  UMR-MP(T) &
  5.2759 &
  12.7835 &
  28.8175 &
  0.2151 &
  21.7829 &
  10.9990 &
  0.0149 &
  66.7192 &
  33.3671 &
  0.3991 &
  0.0501 &
  5.9623 \\ \cdashline{2-14} 
 &
  UMR-MP(SRT) &
  \textbf{2.8805} &
  4.6331 &
  13.4384 &
  0.3746 &
  7.3802 &
  \textbf{3.8774} &
  0.0097 &
  30.1516 &
  \textbf{15.0807} &
  0.5013 &
  0.0383 &
  8.6322 \\ \cline{2-14} 
 &
  MR-MP &
  3.5060 &
  6.8842 &
  17.3479 &
  4.6039 &
  2.8674 &
  3.7357 &
  0.5465 &
  8.6723 &
  \textbf{4.6094} &
  0.6769 &
  0.1097 &
  7.9614 \\ \cdashline{2-14} 
 &
  MR-MP(T) &
  4.2947 &
  7.9124 &
  16.3772 &
  94.0982 &
  0.0064 &
  47.0523 &
  61.3860 &
  0.6354 &
  31.0107 &
  0.8082 &
  0.4716 &
  6.6455 \\ \cdashline{2-14} 
\multirow{-7}{*}{
\rotatebox[origin=c]{90}{MobileFaceNet}
} &
  MR-MP(SRT) &
  \textbf{3.1866} &
  5.6166 &
  13.5290 &
  3.1906 &
  3.1867 &
  \textbf{3.1886} &
  0.2658 &
  9.4802 &
  4.8730 &
  0.6636 &
  0.0837 &
  8.0905 \\ \hline
\end{tabular}%
}
\caption{The achieved verification performance of different experimental settings by ResNet-100, ResNet-50 and MobileFaceNet models along with EUM trained with triplet loss and EUM trained with \gls{srt} loss. The result is reported on MFR dataset. The FMR100\_Th\textsuperscript{UMR-UMP} are equal to 0.2307, 0.2652 and 0.3246 for ResNet-100, ResNet-50 and MobileFaceNet, respectively. The FMR1000\_Th\textsuperscript{UMR-UMP} are equal to 0.3482, 0.3926 and 0.4476 for ResNet-100, ResNet-50 and MobileFaceNet, respectively.
The lowest EER and the lowest average error of FMR100 and FMR1000 at the defined threshold for each of the evaluation cases are marked in bold.
One can notice the significant improvement in the verification performance induced by our proposed approach (\gls{srt}) in most evaluation cases.}
\label{tab:mfr}
\end{table*}

\section{Result} 
\label{sec:result}
In this section, we present and discuss our achieved results.
First of all, we experimentally present the negative impact of wearing a face mask on face recognition performance. 
Then, we present and discuss the impact of our \gls{eum} trained with \gls{srt} on enhancing the separability between the genuine and imposter comparison scores. 
Then, we present the gain in the masked face verification performance achieved by our proposed \gls{eum} trained with \gls{srt} on the collaborative and in-the-wild masked face recognition.  
Finally, we present an ablation study on \gls{srt} to experimentally support our theoretical motivation behind the  \gls{srt} loss by comparing its performance with the triplet loss. 
\subsection{Impact of Masked Face on the Face Recognition Verification Performance}
Tables \ref{tab:mfr}, \ref{tab:mrf2}, \ref{tab:lfw}, and \ref{tab:ijbc} present a comparison between the baseline evaluation where reference and probe are unmasked (UMR-UMP), the case where only the probe is masked (UMR-MP), and the case where reference and probe are masked (MR-MP).
On UMR-UMP case, the considered face recognition models, ResNet-100,  ResNet-50, and MobileFaceNet, achieve a very high verification performance. This is demonstrated by scoring 0.0\%, 0.0\% and 0.0\% EER on the MFR dataset (Table \ref{tab:mfr}), 0.0\%, 0.0\% and 0.0106\% on the MRF2 dataset (Table \ref{tab:mrf2}), 0.2660\%, 0.3333\% and 0.6333\% EER on the LFW (Table \ref{tab:lfw}) and 1.5340\%, 1.6881\% and 2.2396\% EER on IJB-C dataset (Table \ref{tab:ijbc}), respectively, by model ResNet-100, ResNet-50 and MobileFaceNet. 

The verification performances of the considered models are substantially degraded when \textcolor{revision}{ evaluated on real and synthetically generated masked face images.}
This is indicated by the degradation in verification performance measures and FDR values, in comparison to the case where probe and reference are unmasked.
\textcolor{revision}{MobileFaceNet achieved lower verification performance on MR-MP than UMR-MP evaluation setting, as seen in Tables \ref{tab:mfr}, \ref{tab:mrf2}, \ref{tab:lfw}, and \ref{tab:ijbc}. Furthermore, ResNet-50 achieved lower verification performance in the MR-MP than the UMR-MP setting when it is evaluated on MFR, MRF2, and IJB-C datasets. For example, on the MFR dataset, the achieved EER by ResNet-50 model is 1.2492\% (UMR-MP). This error rate is raised to 1.2963\% for the MR-MP evaluation setting, as seen in Table \ref{tab:mfr}.
On LFW, the ResNet-50 model achieved very close performance for the MR-MP and the UMR-MP evaluation setting.
In this case, the achieved EER by ResNet-50 are 1.4667\% for the UMR-MP evaluation setting and 1.3667\% for the MR-MP evaluation setting.  
Furthermore, ResNet-100 achieved lower verification performance for the MR-MP evaluation setting than the UMR-MP evaluation setting when it is evaluated on MRF2 and IJB-C datasets. On LFW and MFR, the ResNet-100 model achieved very close performance for the MR-MP and the UMR-MP evaluation settings.
When the FMR and FNMR measures are calculated based on FMR100\_Th\textsuperscript{UMR-UMP}, the achieved 
FMR and FNMR are higher on the MR-MP than the UMR-MP case in most of the settings. When the threshold is set to FMR1000\_Th\textsuperscript{UMR-UMP}, the achieved FMR and FNMR are lower when both reference and probe are masked (MR-MP) than in the case where only probes are masked (UMR-MP) in most of the evaluation settings. Also, one can be noticed that wearing a face mask (UMR-MP and MR-MP cases) has a higher effect on the FNMR than FMR when these measures are calculated based on FMR100\_Th\textsuperscript{UMR-UMP} or FMR1000\_Th\textsuperscript{UMR-UMP}. }
These results are also supported by the G-mean, I-mean, and FDR shown in Tables \ref{tab:mfr}, \ref{tab:mrf2}, \ref{tab:lfw} and \ref{tab:ijbc}. 

We also make four general observations: 1) The compact model, MobileFaceNet, achieved lower verification performance than the  ResNet-100 and ResNet-50 model. One of the reasons for this performance degradation might be due to the smaller embedding size of MobileFaceNet (128-D), in comparison to the embedding size of 512-D in ResNet-100 and ResNet-50.
Moreover, the size of the MobileFaceNet network (1m parameters) is extremely smaller than ResNet-100 (65m parameters) and ResNet-50 (36m parameters), which might affect the generalization ability of the MobileFaceNet model.
2) The considered models achieved lower performance when evaluated on the MRF2 dataset than the case when evaluated on the MFR dataset. This result was expected as the images in the MRF2 dataset are crawled from the internet with high variations in facial expression, pose, illumination. On the other hand, the images in the MFR dataset are collected in a collaborative environment.  3) The considered models achieved lower performance on LFW and IJB-C datasets in comparison to MFR and MRF2 as they are larger scale.
The considered models achieved lower performance when evaluated on IJB-C than the case when evaluated on LFW.
This result was expected as the evaluation protocol of LFW is simpler than the IJB-C,  and the IJB-C has shown to be more challenging than LFW in multiple studies \cite{DBLP:conf/cvpr/DengGXZ19,DBLP:conf/fgr/CaoSXPZ18}. 4) The considered models achieved relatively higher G-mean scores on the UMR-MP than the MR-MP experimental setting. This indicates a higher similarity between genuine pairs in the MR-MP than the UMR-MP. However, the achieved verification performances by UMR-MP cases on most of the evaluated datasets are higher than the achieved ones by MR-MP.  One of the contributing factors for the difference in the performance is that the imposter distribution is shifted more toward genuine distribution in the MR-MP cases than the UMR-MP ones, i.e. masked face pairs are more similar (in comparison to unmasked-masked pairs) even if the identities are different. This statement can be clearly observed from the achieved I-mean values shown in Tables \ref{tab:mfr}, \ref{tab:mrf2}, \ref{tab:lfw}, and \ref{tab:ijbc}.  This shifting in imposter distribution strongly affects the verification performance of the considered models. 

To summarize, wearing a face mask has a negative effect on face recognition performance. This observation is experimentally proved by evaluating the verification performance of three face recognition models, ResNet-100, ResNet-50, and MobileFaceNet, on two real masked datasets (MFR and MRF2) and two synthetically generated masked face datasets (LFW and IJB-C). This result supports and complements the previous findings in the studies in \cite{DBLP:conf/biosig/DamerGCBKK20,DHS-Rally-2020,ngan2020ongoing,ngan2020ongoingpost} evaluating the impact of wearing a mask on face recognition performance.   
\subsection{Impact of our EUM with \gls{srt} on the Separability between Genuine and Imposter Comparison Scores}
The proposed approach significantly enhanced the separability between the genuine and imposter comparison scores in the considered face recognition models and both evaluated datasets. This improvement can be seen in the increase in the FDR separability measure achieved by our proposed \gls{eum} trained with\gls{srt} in comparison to the achieved FDR measures by the considered face recognition models, as shown in Table\ref{tab:mfr}, \ref{tab:mrf2}, \ref{tab:lfw}, and \ref{tab:ijbc}.
This indicates a general improvement in the verification performance of face recognition and thus enhancing the general trust in the verification decision.
For example, when the 
ResNet-50 model is evaluated on the MFR dataset and the probe is masked, the FDR increases from 12.6189 (UMR-MP) to 13.4045 (UMR-MP(SRT)) using our proposed approach, as shown in Table \ref{tab:mfr}. 
Similar observations can be made when the evaluation dataset is synthetically masked. For example, when ResNet-100 is evaluated on the synthetically generated masked face of IJB-C, the FDR increases from 9.7516 (UMR-MP) to 9.9005 (UMR-MP(SRT)) using our proposed approach, as shown in Table \ref{tab:ijbc}. 
This improvement in the separability between the genuine and the imposter samples by our proposed approach is achieved in most of the evaluation settings, where the FDR increased in 20 out of 24 experimental settings.

\subsection{Impact of our EUM with SRT Solution on the Collaborative Masked Face Recognition}
\label{sec:mfr}
When the considered models are evaluated on the MFR dataset, it can be observed that our proposed approach significantly enhanced the masked face verification performance, as shown in Table \ref{tab:mfr}.
The achieved EER by ResNet-100 is 0.8912\% on the UMR-MP case. This error is reduced to 0.7702\% using our approach (UMR-MP(SRT)). 
The achieved EER by the ResNet-100 is 0.8014\% on MR-MP experimental setting. The achieved EER using our approach on top of the ResNet-100 is 0.8270\%  (MR-MP(SRT)). However, this is the only case that we did not observe improvement in EER when the considered models are evaluated on the MFR dataset.
The achieved EER by ResNet-50 model is 1.2492\% based on UMR-MP experimental setting. This error rate is decreased to 0.9611\% by our proposed approach (UMR-MP(SRT)) indicating a clear improvement in the verification performance induced by our proposed approach, as shown in Table \ref{tab:mfr}.  
A similar enhancement in the verification performance is observed by our approach for the MR-MP setting. In this case, the EER is decreased from 1.2963\% (MR-MP) to 1.1207\% (MR-MP(SRT)).
The achieved EER by the MobileFaceNet model is 3.4939\% (UMR-MP). This error is reduced to 2.8805\% using our proposed approach (UMR-MP(SRT)). Considering the MR-MP setting, the EER is decreased from 3.506\% (MR-MP) to 3.1866\% (MR-MP(SRT)) by our approach.
The improvement in the masked face recognition verification performance is also noticeable from the improvement in FMR100 and FMR1000 measures. 
When the considered models are evaluated on masked data (UMR-MP and MR-MP) based on  FMR100\_Th\textsuperscript{UMR-MP}, the average of FMR and FNMR was significantly reduced by our proposed approach in all evaluation settings (UMR-MP(SRT) and MR-MP(SRT)), as shown in Table \ref{tab:mfr}. When the operation threshold is calculated at FMR1000 (FMR1000\_Th\textsuperscript{UMR-MP}), a significant reduction in the average of FMR and FNMR with our proposed approach is notable in most evaluation settings. This result is also supported by ROC curves shown in Figures \ref{fig:roc_r100_mfr}, \ref{fig:roc_r50_mfr} and \ref{fig:roc_m_mfr}.

\begin{table*}[ht!]
\centering
\resizebox{\textwidth}{!}{%
\begin{tabular}{|l|l|l|l|l|l|l|l|l|l|l|l|l|l|}
\hline
 &
   &
  \cellcolor[HTML]{FFFFFF}{\color[HTML]{000000} } &
   &
   &
  \multicolumn{3}{l|}{FMR100\_Th\textsuperscript{UMR-UMP}} &
  \multicolumn{3}{l|}{FMR1000\_Th\textsuperscript{UMR-UMP}} &
   &
   &
   \\ \cline{6-11}
\multirow{-2}{*}{MRF2} &
  \multirow{-2}{*}{Setting} &
  \multirow{-2}{*}{\cellcolor[HTML]{FFFFFF}{\color[HTML]{000000} EER\%}} &
  \multirow{-2}{*}{FMR100\%} &
  \multirow{-2}{*}{FMR1000\%} &
  FMR\% &
  FNMR\% &
  Avg.\% &
  FMR\% &
  FNMR\% &
  Avg.\% &
  \multirow{-2}{*}{G-mean} &
  \multirow{-2}{*}{I-mean} &
  \multirow{-2}{*}{FDR} \\ \hline
 &
  UMR-UMP &
  0.0000 &
  0.0000 &
  0.0000 &
  1.0000 &
  0.0000 &
  0.5000 &
  0.1000 &
  0.0000 &
  0.0500 &
  0.7605 &
  0.0019 &
  46.4218 \\ \cline{2-14} 
 &
  UMR-MP &
  4.0515 &
  6.7568 &
  7.0946 &
  0.9079 &
  6.7568 &
  3.8323 &
  0.1127 &
  7.0946 &
  \textbf{3.6036} &
  0.4454 &
  -0.0000 &
  9.3458 \\ \cdashline{2-14} 
 &
  UMR-MP(T) &
  4.0515 &
  6.7568 &
  9.4595 &
  0.7820 &
  6.7568 &
  3.7694 &
  0.0530 &
  11.1486 &
  5.6008 &
  0.3677 &
  -0.0012 &
  8.3377 \\ \cdashline{2-14} 
 &
  UMR-MP(SRT) &
  \textbf{3.3757} &
  5.4054 &
  7.0946 &
  0.9145 &
  5.7432 &
  \textbf{3.3289} &
  0.1127 &
  7.0946 &
  \textbf{3.6036} &
  0.4587 &
  -0.0003 &
  9.8264 \\ \cline{2-14} 
 &
  MR-MP &
  3.7522 &
  3.7559 &
  8.4507 &
  4.3648 &
  3.4429 &
  3.9039 &
  1.0079 &
  3.7559 &
  \textbf{2.3819} &
  0.6757 &
  0.0183 &
  6.4714 \\ \cdashline{2-14} 
 &
  MR-MP(T) &
  4.3817 &
  9.0767 &
  21.5962 &
  20.6247 &
  2.5039 &
  11.5643 &
  9.3461 &
  3.1299 &
  6.2380 &
  0.6947 &
  0.0834 &
  5.8089 \\ \cdashline{2-14} 
\multirow{-7}{*}{\rotatebox[origin=c]{90}{ResNet-100}} &
  MR-MP(SRT) &
  \textbf{3.4416} &
  4.3818 &
  8.4507 &
  3.8651 &
  3.1299 &
  \textbf{3.4975} &
  0.8247 &
  4.3818 &
  2.6033 &
  0.6738 &
  0.0099 &
  6.4496 \\ \hline \hline
 &
  UMR-UMP &
  0.0000 &
  0.0000 &
  0.0000 &
  1.0000 &
  0.0000 &
  0.5000 &
  0.1000 &
  0.0000 &
  0.0500 &
  0.7477 &
  0.0038 &
  37.9345 \\ \cline{2-14} 
 &
  UMR-MP &
  \textbf{4.3895} &
  6.7568 &
  10.4730 &
  0.7025 &
  8.4459 &
  4.5742 &
  0.0795 &
  10.8108 &
  5.4452 &
  0.4263 &
  0.0005 &
  8.2432 \\ \cdashline{2-14} 
 &
  UMR-MP(T) &
  6.4169 &
  7.7703 &
  12.1622 &
  0.4241 &
  8.7838 &
  4.6040 &
  0.0000 &
  17.9054 &
  8.9527 &
  0.3567 &
  -0.0066 &
  6.8853 \\ \cdashline{2-14} 
 &
  UMR-MP(SRT) &
  4.7274 &
  7.4324 &
  9.4595 &
  0.8748 &
  7.4324 &
  \textbf{4.1536} &
  0.1193 &
  9.1216 &
  \textbf{4.6205} &
  0.4553 &
  0.0014 &
  8.4507 \\ \cline{2-14} 
 &
  MR-MP &
  6.8831 &
  10.0156 &
  13.7715 &
  4.2316 &
  7.8247 &
  6.0281 &
  1.1662 &
  9.7027 &
  5.4344 &
  0.6496 &
  0.0301 &
  4.7924 \\ \cdashline{2-14} 
 &
  MR-MP(T) &
  6.8831 &
  9.7027 &
  14.0845 &
  97.8759 &
  0.0000 &
  48.9379 &
  90.7622 &
  0.0000 &
  45.3811 &
  0.7759 &
  0.3663 &
  4.8791 \\ \cdashline{2-14} 
\multirow{-7}{*}{\rotatebox[origin=c]{90}{ResNet-50}} &
  MR-MP(SRT) &
  \textbf{6.2578} &
  9.0767 &
  11.8936 &
  2.9738 &
  8.1377 &
  \textbf{5.5557} &
  0.8413 &
  9.3897 &
  \textbf{5.1155} &
  0.6488 &
  0.0144 &
  4.9381 \\ \hline \hline
 &
  UMR-UMP &
  0.0106 &
  0.0000 &
  0.0000 &
  1.0000 &
  0.0000 &
  0.5000 &
  0.1000 &
  0.0000 &
  0.0500 &
  0.7318 &
  0.0078 &
  26.4276 \\ \cline{2-14} 
 &
  UMR-MP &
  6.4169 &
  16.8919 &
  24.3243 &
  0.9874 &
  16.8919 &
  8.9397 &
  0.0663 &
  27.3649 &
  13.7156 &
  0.3803 &
  -0.0019 &
  4.6457 \\ \cdashline{2-14} 
 &
  UMR-MP(T) &
  7.7685 &
  15.8784 &
  34.4595 &
  0.6759 &
  18.9189 &
  9.7974 &
  0.0596 &
  37.1622 &
  18.6109 &
  0.3304 &
  -0.0027 &
  4.2067 \\ \cdashline{2-14} 
 &
  UMR-MP(SRT) &
  \textbf{6.079} &
  12.5000 &
  21.9595 &
  0.9675 &
  13.1757 &
  \textbf{7.0716} &
  0.0928 &
  22.2973 &
  \textbf{11.1950} &
  0.4157 &
  -0.0018 &
  5.2918 \\ \cline{2-14} 
 &
  MR-MP &
  8.4777 &
  18.1534 &
  28.7950 &
  6.5056 &
  10.3286 &
  8.4171 &
  1.9908 &
  14.0845 &
  8.0377 &
  0.6087 &
  0.0509 &
  3.2505 \\ \cdashline{2-14} 
 &
  MR-MP(T) &
  8.7634 &
  17.5274 &
  26.2911 &
  95.9683 &
  0.0000 &
  47.9842 &
  84.9896 &
  0.0000 &
  42.4948 &
  0.7638 &
  0.3966 &
  3.5408 \\ \cdashline{2-14} 
\multirow{-7}{*}{\rotatebox[origin=c]{90}{MobileFaceNet}} &
  MR-MP(SRT) &
  \textbf{7.8232} &
  15.0235 &
  22.5352 &
  3.9733 &
  9.0767 &
  \textbf{6.525} &
  1.1745 &
  14.3975 &
  \textbf{7.7860} &
  0.6087 &
  0.0241 &
  3.5815 \\ \hline
\end{tabular}%
}
\caption{The achieved verification performance of different experimental settings by ResNet-100, ResNet-50 and MobileFaceNet models along with EUM trained with triplet loss and EUM trained with \gls{srt} loss. The result is reported using MRF2 dataset. The FMR100\_Th\textsuperscript{UMR-UMP} are equal to 0.1711, 0.2038 and 0.2351 for ResNet-100, ResNet-50 and MobileFaceNet, respectively. The FMR1000\_Th\textsuperscript{UMR-UMP} are equal to 0.2316, 0.2639 and 0.3041 for ResNet-100, ResNet-50 and MobileFaceNet, respectively.
The lowest EER and the lowest average error of FMR100 and FMR1000 at the defined threshold for each of the evaluation cases and each of the evaluated models are marked in bold.
One can notice the significant improvement in the verification performance induced by our proposed approach (\gls{srt}) in most evaluation cases.}
\label{tab:mrf2}
\end{table*}

\subsection{Impact of our EUM with SRT on in-the-wild Masked Face Recognition}
\label{sec:mrf2}
The achieved evaluation results on the MRF2 dataset by ResNet-100, ResNet-50, and MobileFaceNet models are presented in Tables \ref{tab:mrf2}.
When probes are masked, the ERR achieved by the ResNet-100 model is reduced from 4.0515\% (UMR-MP) to 3.3757\% by our proposed approach (UMR-MP(SRT)). 
A similar improvement in the verification performance is achieved by our solution (MR-MP(SRT)) in the MR-MP evaluation setting, as shown in Table \ref{tab:mrf2}.

Using masked probes, the achieved  EER by ResNet-50 model is 4.3895\% (UMR-MP). Only in this case, the EER did not improve by our proposed approach (UMR-MP(SRT)). The achieved EER, in this case, by our proposed approach is 4.7274\%. Nonetheless, a notable improvement in the FMR1000 and the FDR separability measures can be observed from the reported result. The increase in FDR points out the possibility that given larger and more representative evaluation data, the consistent enhancement in verification accuracy will be apparent here as well. 
A significant improvement in the verification performance is achieved by our approach when comparing masked probes to masked references. In this case, the achieved EER is decreased from 6.8831\% (MR-MP) to 6.2578\% (MR-MP(SRT)). A similar conclusion can be made from the improvements on the other performance verification measures and the FDR measure.    

Using masked probes, the achieved verification performance by MobileFaceNet is significantly enhanced by our proposed approach (UMR-MP(SRT)).
A similar improvement in the verification performance is achieved on MR-MP(SRT) case as shown in Table \ref{tab:mrf2}. For example, the achieved EER by MobileFaceNet is 8.4777\% on the MR-MP case. This error rate is reduced to 7.8232\% using our proposed approach (MR-MP(SRT)).

Considering the FMR100\_Th\textsuperscript{UMR-UMP} and the FMR1000\_Th\textsuperscript{UMR-UMP}, the achieved FMR and FNMR improved is by our proposed solution (UMR-MP(SRT) and MR-MP(SRT)) 
in most evaluation cases, especially when the considered operation threshold is FMR100\_Th\textsuperscript{UMR-UMP}. This result is also supported by ROC curves shown in Figures \ref{fig:roc_r100_mrf2}, \ref{fig:roc_r50_mrf2} and \ref{fig:roc_m_mrf2}.

\begin{table*}[ht!]
\centering
\resizebox{\textwidth}{!}{%
\begin{tabular}{|l|l|l|l|l|l|l|l|l|l|l|l|l|l|}
\hline
 &
   &
   &
  \cellcolor[HTML]{FFFFFF}{\color[HTML]{000000} } &
   &
  \multicolumn{3}{l|}{FMR100\_Th\textsuperscript{UMR-UMP}} &
  \multicolumn{3}{l|}{FMR1000\_Th\textsuperscript{UMR-UMP}} &
   &
   &
   \\ \cline{6-11}
\multirow{-2}{*}{LFW} &
  \multirow{-2}{*}{Setting} &
  \multirow{-2}{*}{EER\%} &
  \multirow{-2}{*}{\cellcolor[HTML]{FFFFFF}{\color[HTML]{000000} FMR100\%}} &
  \multirow{-2}{*}{FMR1000\%} &
  FMR\% &
  FNMR\% &
  Avg.\% &
  FMR\% &
  FNMR\% &
  Avg.\% &
  \multirow{-2}{*}{G-mean} &
  \multirow{-2}{*}{I-mean} &
  \multirow{-2}{*}{FDR} \\ \hline
 &
  UMR-UMP &
  0.2660 &
  0.2667 &
  0.3333 &
  1.0000 &
  0.2667 &
  0.6333 &
  0.1000 &
  0.3333 &
  0.2167 &
  0.7157 &
  0.0026 &
  33.0630 \\ \cline{2-14} 
 &
  UMR-MP &
  1.0000 &
  0.9667 &
  2.5667 &
  1.0667 &
  0.9667 &
  \textbf{1.0167} &
  0.0667 &
  2.9333 &
  1.5000 &
  0.5220 &
  0.0019 &
  13.1746 \\ \cdashline{2-14} 
 &
  UMR-MP(T) &
  1.7000 &
  2.3667 &
  4.4333 &
  0.9667 &
  2.5333 &
  1.7500 &
  0.0333 &
  5.9667 &
  3.0000 &
  0.4115 &
  0.0029 &
  11.0452 \\ \cdashline{2-14} 
 &
  UMR-MP(SRT) &
  \textbf{0.8667} &
  0.8667 &
  1.6000 &
  1.2667 &
  0.7667 &
  \textbf{1.0167} &
  0.1000 &
  1.7333 &
  \textbf{0.9167} &
  0.5380 &
  0.0024 &
  15.0505 \\ \cline{2-14} 
 &
  MR-MP &
  \textbf{0.9667} &
  0.9667 &
  2.4333 &
  3.1000 &
  0.7000 &
  1.9000 &
  0.5000 &
  1.2000 &
  \textbf{0.8500} &
  0.5996 &
  0.0110 &
  14.2278 \\ \cdashline{2-14} 
 &
  MR-MP(T) &
  1.7333 &
  2.3333 &
  10.1333 &
  19.4667 &
  0.3667 &
  9.9167 &
  6.7667 &
  0.6667 &
  3.7167 &
  0.6290 &
  0.0808 &
  10.8161 \\ \cdashline{2-14} 
\multirow{-7}{*}{\rotatebox[origin=c]{90}{ResNet-100}} &
  MR-MP(SRT) &
  \textbf{0.9667} &
  0.9667 &
  2.0667 &
  3.0000 &
  0.6667 &
  \textbf{1.8333} &
  0.4667 &
  1.5333 &
  1.0000 &
  0.6035 &
  0.0053 &
  14.6018 \\ \hline \hline
 &
  UMR-UMP &
  0.3333 &
  0.3000 &
  0.4000 &
  1.0000 &
  0.3000 &
  0.6500 &
  0.1000 &
  0.4000 &
  0.2500 &
  0.7023 &
  0.0029 &
  26.5107 \\ \cline{2-14} 
 &
  UMR-MP &
  1.4667 &
  1.8333 &
  3.3000 &
  1.0000 &
  1.8333 &
  1.4167 &
  0.1000 &
  3.5667 &
  1.8333 &
  0.5117 &
  0.0014 &
  11.8522 \\ \cdashline{2-14} 
 &
  UMR-MP(T) &
  2.0000 &
  2.7000 &
  4.9667 &
  0.6333 &
  3.3333 &
  1.9833 &
  0.0667 &
  6.6333 &
  3.3500 &
  0.4278 &
  0.0020 &
  10.5553 \\ \cdashline{2-14} 
 &
  UMR-MP(SRT) &
  \textbf{1.1000} &
  1.1333 &
  2.4000 &
  0.9667 &
  1.1333 &
  \textbf{1.0500} &
  0.2000 &
  2.2000 &
  \textbf{1.2000} &
  0.5427 &
  0.0016 &
  14.5079 \\ \cline{2-14} 
 &
  MR-MP &
  1.3667 &
  1.7333 &
  4.7333 &
  3.0000 &
  0.8333 &
  1.9167 &
  0.9000 &
  1.9333 &
  1.4167 &
  0.5893 &
  0.0158 &
  12.2339 \\ \cdashline{2-14} 
 &
  MR-MP(T) &
  2.0333 &
  2.9667 &
  7.2000 &
  10.8667 &
  0.7667 &
  5.8167 &
  4.0333 &
  1.5333 &
  2.7833 &
  0.6256 &
  0.0525 &
  10.2560 \\ \cdashline{2-14} 
\multirow{-7}{*}{\rotatebox[origin=c]{90}{ResNet-50}} &
  MR-MP(SRT) &
  \textbf{1.2333} &
  1.4333 &
  2.9667 &
  2.2333 &
  0.9333 &
  \textbf{1.5833} &
  0.6333 &
  1.5333 &
  \textbf{1.0833} &
  0.6051 &
  0.0053 &
  13.4416 \\ \hline \hline
 &
  UMR-UMP &
  0.6333 &
  0.6000 &
  1.3000 &
  1.0000 &
  0.6000 &
  0.8000 &
  0.1000 &
  1.3000 &
  0.7000 &
  0.6742 &
  0.0051 &
  18.2460 \\ \cline{2-14} 
 &
  UMR-MP &
  3.2333 &
  5.9333 &
  12.0333 &
  0.7667 &
  6.7333 &
  3.7500 &
  0.0000 &
  18.2667 &
  9.1333 &
  0.4641 &
  -0.0011 &
  7.5840 \\ \cdashline{2-14} 
 &
  UMR-MP(T) &
  3.6667 &
  7.1333 &
  17.6667 &
  0.6000 &
  8.7667 &
  4.6833 &
  0.0000 &
  27.4333 &
  13.7167 &
  0.4023 &
  0.0013 &
  7.2341 \\ \cdashline{2-14} 
 &
  UMR-MP(SRT) &
  \textbf{1.8667} &
  2.4667 &
  8.1333 &
  0.8333 &
  2.8667 &
  \textbf{1.8500} &
  0.1000 &
  9.3667 &
  \textbf{4.7333} &
  0.5144 &
  0.0006 &
  10.2266 \\ \cline{2-14} 
 &
  MR-MP &
  3.3333 &
  6.4667 &
  17.9000 &
  5.7667 &
  2.6333 &
  4.2000 &
  0.8333 &
  7.1333 &
  3.9833 &
  0.5688 &
  0.0505 &
  7.7096 \\ \cdashline{2-14} 
 &
  MR-MP(T) &
  3.0667 &
  5.2000 &
  13.6333 &
  93.9000 &
  0.0000 &
  46.9500 &
  72.1333 &
  0.0667 &
  36.1000 &
  0.7495 &
  0.3970 &
  7.7594 \\ \cdashline{2-14} 
\multirow{-7}{*}{\rotatebox[origin=c]{90}{MobileFaceNet}} &
  MR-MP(SRT) &
  \textbf{2.2667} &
  3.5333 &
  11.1000 &
  2.3000 &
  2.2333 &
  \textbf{2.2667} &
  0.4667 &
  5.9667 &
  \textbf{3.2167} &
  0.5872 &
  0.0091 &
  9.6183 \\ \hline
\end{tabular}%
}
\caption{ The achieved verification performance of different experimental settings by ResNet-100, ResNet-50, and MobileFaceNet models along with EUM trained with triplet loss and EUM trained with \gls{srt} loss. The result is reported using synthetically generated masked faces of the LFW dataset. The FMR100\_Th\textsuperscript{UMR-UMP} are equal to 0.1736, 0.2052 and 0.2449  for ResNet-100, ResNet-50 and MobileFaceNet, respectively. The FMR1000\_Th\textsuperscript{UMR-UMP} are equal to 0.2451, 0.2617 and 0.3450 are  for ResNet-100, ResNet-50 and MobileFaceNet, respectively.
The lowest EER and the lowest average error of FMR100 and FMR1000 at the defined threshold are marked in bold.
It can be noticed the significant improvement in the verification performance induced by our proposed approach (\gls{srt}) in most evaluation cases.}
\label{tab:lfw}
\vspace{-4mm}
\end{table*}

\subsection{Impact of our EUM with SRT on Simulated Masked Face Recognition}
In addition to the evaluation of the real masked face dataset presented in Section \ref{sec:mfr} and \ref{sec:mrf2}, we evaluated our proposed solution on two large synthetically generated masked faces datasets: LFW and IJB-C.
The achieved verification performance on the synthetically generated masked face of LFW is presented in Table \ref{tab:lfw}.  The improvement in verification performance induced by our proposed solution on the synthetic masked face of LFW is observable for all evaluation cases.

Table \ref{tab:ijbc} presents the achieved verification performance by the considered models on the synthetically generated masked face of IJB-C. When the reference and the probes are synthetically masked, the achieved EER by ResNet-100 is 2.7356\% (MR-MP).  Only in this case for synthetically masked face dataset, the EER did not improve by our proposed approach, where the EER achieved by our approach is 2.9197\% (MR-MP(SRT)). However, when the operation threshold is set to FMR100\_Th\textsuperscript{UMR-UMP}, a notable reduction in the average of FMR and FNMR can be observed for all evaluation cases. These reported results on synthetically generated masked face datasets support our achievement on real masked face datasets. Also, it points out the competence of our proposed solution in improving the masked face verification performance. A similar observation can be noticed in the ROC curves in Figures \ref{fig:roc_r100_lfw}, \ref{fig:roc_r50_lfw}, \ref{fig:roc_m_lfw}, \ref{fig:roc_r100_ijbc}, \ref{fig:roc_r50_ijbc}, \ref{fig:roc_m_ijbc}.  

\subsection{Ablation Study on Self-restrained Triplet Loss}
In this subsection, we experimentally prove and theoretically discuss the advantage of our proposed \gls{srt} solution over the common naive triplet loss.
We explore first the validity of training the EUM model with triplet loss using masked face datasets. It is noticeable that training EUM with naive triplet is inefficient for learning from masked face embedding as presented in in Tables \ref{tab:mfr}, \ref{tab:mrf2}, \ref{tab:lfw} and \ref{tab:ijbc}.
For example, when the probe is masked, the achieved EER by \gls{eum} with triplet loss on top of ResNet-50 is 1.9789\% (UMR-MP(T)), in comparison to 0.9611\% EER achieved by \gls{eum} with our\ \gls{srt} (UMR-MP(SRT)), as shown in Table \ref{tab:mfr}. 
It is crucial for learning with triplet loss that the input triplet violate the condition $d(f(x^a_{i}),f(x^n_{i}))> d(f(x^a_{i}),f(x^p_{i}))+ m$. Thus, the model can learn to minimize the distance between the genuine pairs and maximize the distance between the imposter pairs.  When the previous condition is not violated, the loss value will be close to zero and the model will not be able to further optimize the distances of the genuine pairs and imposter pairs. This motivated our \gls{srt} solution.

Given that our proposed \gls{eum} solution is built on top of a pre-trained face recognition model, the feature embeddings of the genuine pairs are similar (to a large degree), and the ones of imposter pairs are dissimilar.
However, this similarity is affected (to some degree) when the faces are masked. 
The learning objective of our approach is to reduce this effect.
This statement can be observed from the achieved results presented in Tables \ref{tab:mfr}, \ref{tab:mrf2}, \ref{tab:lfw} and \ref{tab:ijbc}. 
For example,  using the MFR dataset,
the achieved G-mean and I-mean by ResNet-50 is 0.8538 and 0.0349 (UMR-UMP), respectively.  When the probe is masked (UMR-MP), the achieved G-mean and I-mean shift to 0.5254 and 0.0251, respectively, as shown in Table \ref{tab:mfr}. The shifting in the G-mean points out that the similarity between the genuine pairs is reduced (to some degree) when the probe is masked.
Training \gls{eum} with naive triplet loss requires selecting a triplet of embeddings that violated the triplet condition.
As we discussed earlier, the masked anchor is similar (to some degree) to the positive (unmasked embedding), and it is dissimilar (to some degree) from the negative. Therefore, finding triplets that violate the triplet condition is not trivia. Also, it could not be possible for many triplets in the training dataset.
This explains the poor result achieved when the EUM model is trained with triplet loss, as there are only a few triplets violating the triplet loss condition. One can assume that using a larger margin value allows the EUM model to further optimize the distance between genuine pairs and imposter pairs, as the triplet condition can be violated by increasing the margin value. 
However, by increasing the margin value, we increase the upper bound of the loss function.
Thus, we ignore the fact that the distance between imposter pairs is sufficient with respect to the distance between genuine pairs in the embedding space. For example, using unmasked data, the mean of the imposter scores achieved by ResNet-50 on the MFR dataset is 0.0349.  When the probe is masked, the mean value of the imposter scores is 0.0251, as shown in Table \ref{tab:mfr}.
Therefore, any further optimization on the distance between the imposter pairs will affect the discriminative features learned by the base face recognition model. Also, there is no restriction in the learning process ensured that the model will maintain the distance between the imposter pairs.
Alternatively, training the EUM model with our \gls{srt} loss achieved significant improvement in minimizing the distance between the genuine pairs. Simultaneously, it maintains the distance between the imposter pairs to be closer to the one learned by the base face recognition model. 
It is noticeable from the reported result that the I-mean achieved by our \gls{srt} is closer to the I-mean achieved when the model is evaluated on unmasked data, in comparison to the one achieved by naive triplet loss, as shown in Tables \ref{tab:mfr}, \ref{tab:mrf2}, \ref{tab:lfw} and \ref{tab:ijbc}.
The achieved result points out the efficiency of our proposed \gls{eum} trained with \gls{srt} in improving the masked face recognition, in comparison to the considered face recondition models. Also, it supported our theoretical motivation behind \gls{srt} where training the \gls{eum} with \gls{srt} significantly outperformed the \gls{eum} trained with naive triplet loss.

The proposed solution is designed and trained to manipulate masked face embedding and not to manipulate unmasked one. Based on this workflow, our EUM solution will not be used on unmasked faces. This is based on the assumption that the existence of the mask is known, e.g., by the automatic detection of wearing a face mask that can be relatively easily detected where most of mask face detection methods proposed in the literature achieved very high accuracy in detecting masked face (more than 99\% \cite{loey2021hybrid}). 
Despite the fact that our workflow does not assume processing unmasked faces, and for the sake of experiment completeness, we apply our solution on UMR-UMP cases. 
The achieved results showed slight degradation in face verification performance in a number of the experimental settings. However, this result was expected as the proposed solution is designed and trained to operate on masked face embedding rather than processing an unmasked face embedding. In the following, we present the achieved results by our proposed approach when it is applied to the UMR-UMP case. When ResNet-100 and ResNet-50 are evaluated on the MFR and MRF2 datasets, and the unmasked face embeddings (UMR-UMP) are processed by \gls{eum} with the \gls{srt} solution, the achieved EER and FMR100 are 0.0\% and 0.0\%, respectively. When MobileFaceNet is evaluated on the MFR or MRF2 datasets and the unmasked face embedding (UMR-UMP) are processed by \gls{eum} with the \gls{srt} solution, the verification performances are slightly degraded. In this case, the EER increases from 0.0\% to 0.0112\% EER, when MobileFaceNet is evaluated on the MFR dataset. When MobileFaceNet is evaluated on the MRF2 dataset, the EER value increases from 0.0106\% to 0.2124\%. The achieved FMR100, in this case, is 0.0\%.
When the considered models are evaluated on the LFW and the unmasked face embeddings (UMR-UMP) are processed by \gls{eum} with the \gls{srt} solution, the verification performances obtained by the considered models slightly deteriorate. In this case, the EER and the FMR100 by the ResNet-100 model decrease from 0.2660\% and 0.2667\% to 0.3\% and 0.2667\%, respectively. When the considered model is ResNet-50, the achieved EER and FMR100 are degraded from 0.3333\% and 0.3000\% to 0.5333\% and 0.5000\%, respectively. For the MobileFaceNet model, the achieved EER and FMR100 are degraded from 0.6333\% and 0.6000\% to 1.1667\% and 1.2000\%, respectively. By applying our approach on UMR-UMP cases of the IJB-C dataset, the achieved verification performance is degraded from 1.5340\% to 1.5595\% EER and from 1.6362\% to 1.7027\% FMR on top of the ResNet-100 model. For the ResNet-50 model, the achieved EER and FMR100 are degraded from 1.6881\% to 2.0709\% EER and from 1.8663\% to 2.4857\% FMR.  For the MobileFaceNet model, the achieved EER and FMR are degraded from 2.396\% to 2.8379\% EER and from 2.7918\% to 4.1417\%.
This performance trend in the UMR-UMP setting is expected as processing unmasked face embedding by \gls{eum} with \gls{srt} is not the aim of our proposed solution and do not match its operational concept, where unmasked faces will not be processed by the \gls{eum}. The conducted experiments are thus only included for the sake of experiment completeness.  
\subsection{Discussion}
In summary, the reported results in this paper illustrate how the verification performance of current face recognition models proposed in the literature is affected by wearing a face mask and how this can be improved by learning to process the masked face embedding to behave more similarly to an embedding from an unmasked face. This has been demonstrated through extensive experimental evaluations of three face recognition models and four masked face datasets. The evaluation datasets include two real masked datasets captured under different scenarios: in the wild (MRF2) and collaborative (MFR), and two synthetically generated masked faces of large-scale datasets: LFW and IJB-C. 
We have also theoretically and experimentally demonstrated the competence of our proposed \gls{eum} together with \gls{srt} in reducing the negative influence of the masked face on the face recognition performance. The competence of our solution in improving the masked face verification performance has been demonstrated on real masked datasets captured under different scenarios (in the wild and collaborative) and on synthetically generated masked faces of large-scale datasets.

From research to industry perspective, the developers of commercial face recognition solutions could use our proposed concept to improve the performance of their algorithms when processing masked face images.
Many commercial face recognition solutions produce face templates to enable template storage instead of images in large-scale datasets. The advantages of storing face templates are to enable faster identification searches and matching, by avoiding the re-generation of embeddings in every search. As our solution operates on embedding space, the commercial models can benefit from our solution to improve the performance of their algorithms when facing a masked face image. Examples of such commercial solutions are Neurotechnology \cite{neurotechnology} and Cognitec \cite{cognitec} (achieved high accuracies in NIST Ongoing Face Recognition Vendor Test (FRVT)\cite{grother2018ongoing}). Such solutions produce face templates to populate the biometric datasets to enable efficient biometric searches.

\begin{table*}[ht!]
\centering
\resizebox{\textwidth}{!}{%
\begin{tabular}{|l|l|l|l|l|l|l|l|l|l|l|l|l|l|}
\hline
 &
   &
   &
   &
  \cellcolor[HTML]{FFFFFF}{\color[HTML]{000000} } &
  \multicolumn{3}{l|}{FMR100\_Th\textsuperscript{UMR-UMP}} &
  \multicolumn{3}{l|}{FMR1000\_Th\textsuperscript{UMR-UMP}} &
   &
   &
   \\ \cline{6-11}
\multirow{-2}{*}{IJB-C} &
  \multirow{-2}{*}{Setting} &
  \multirow{-2}{*}{EER\%} &
  \multirow{-2}{*}{FMR100\%} &
  \multirow{-2}{*}{\cellcolor[HTML]{FFFFFF}{\color[HTML]{000000} FMR1000\%}} &
  FMR\% &
  FNMR\% &
  Avg.\% &
  FMR\% &
  FNMT &
  Avg.\% &
  \multirow{-2}{*}{G-mean} &
  \multirow{-2}{*}{I-mean} &
  \multirow{-2}{*}{FDR} \\ \hline
 &
  UMR-UMP &
  1.5340 &
  1.6362 &
  2.4799 &
  1.0000 &
  1.6362 &
  1.3181 &
  0.1000 &
  2.4799 &
  1.2900 &
  0.7460 &
  0.0034 &
  15.7436 \\ \cline{2-14} 
 &
  UMR-MP &
  2.6026 &
  3.3492 &
  5.8751 &
  1.0684 &
  3.3032 &
  2.1858 &
  0.1133 &
  5.6502 &
  2.8817 &
  0.5593 &
  0.0050 &
  9.7516 \\ \cdashline{2-14} 
 &
  UMR-MP(T) &
  5.0724 &
  8.3500 &
  13.9643 &
  0.6437 &
  9.2857 &
  4.9647 &
  0.0333 &
  17.0374 &
  8.5353 &
  0.3966 &
  0.0004 &
  6.0168 \\ \cdashline{2-14} 
 &
  UMR-MP(SRT) &
  \textbf{2.5476} &
  3.2520 &
  5.7575 &
  1.0563 &
  3.2214 &
  \textbf{2.1388} &
  0.1112 &
  5.5837 &
  \textbf{2.8474} &
  0.5667 &
  0.0038 &
  9.9005 \\ \cline{2-14} 
 &
  MR-MP &
  \textbf{2.7356} &
  3.7685 &
  6.9643 &
  4.3300 &
  2.3163 &
  3.3232 &
  0.9488 &
  3.7992 &
  \textbf{2.3740} &
  0.6751 &
  0.0228 &
  10.2180 \\ \cdashline{2-14} 
 &
  MR-MP(T) &
  5.2834 &
  14.7415 &
  42.2509 &
  52.8218 &
  0.4909 &
  26.6563 &
  31.5749 &
  1.1403 &
  16.3576 &
  0.7273 &
  0.1981 &
  6.6082 \\ \cdashline{2-14} 
\multirow{-7}{*}{\rotatebox[origin=c]{90}{ResNet-100}} &
  MR-MP(SRT) &
  2.9197 &
  3.9781 &
  7.3631 &
  3.4202 &
  2.7663 &
  \textbf{3.0932} &
  0.6837 &
  4.4588 &
  2.5712 &
  0.6604 &
  0.0129 &
  9.4975 \\ \hline \hline
 &
  UMR-UMP &
  1.6881 &
  1.8663 &
  3.0782 &
  1.0000 &
  1.8663 &
  1.4332 &
  0.1000 &
  3.0782 &
  1.5891 &
  0.7370 &
  0.0061 &
  14.6355 \\ \cline{2-14} 
 &
  UMR-MP &
  2.8634 &
  4.2593 &
  7.9971 &
  1.0257 &
  4.2389 &
  2.6323 &
  0.1045 &
  7.9051 &
  4.0048 &
  0.5505 &
  0.0091 &
  9.1274 \\ \cdashline{2-14} 
 &
  UMR-MP(T) &
  4.9547 &
  8.3602 &
  15.0995 &
  0.5824 &
  9.6436 &
  5.1130 &
  0.0317 &
  19.0878 &
  9.5597 &
  0.4227 &
  0.0005 &
  6.3770 \\ \cdashline{2-14} 
 &
  UMR-MP(SRT) &
  \textbf{2.7221} &
  3.8401 &
  7.4142 &
  1.0675 &
  3.7685 &
  \textbf{2.4180} &
  0.1162 &
  7.1944 &
  \textbf{3.6553} &
  0.5731 &
  0.0061 &
  9.5896 \\ \cline{2-14} 
 &
  MR-MP &
  3.2418 &
  4.9138 &
  10.0680 &
  5.1556 &
  2.6026 &
  3.8791 &
  1.1855 &
  4.6275 &
  \textbf{2.9065} &
  0.6698 &
  0.0395 &
  9.2267 \\ \cdashline{2-14} 
 &
  MR-MP(T) &
  4.8065 &
  10.6202 &
  30.2398 &
  28.2029 &
  1.1556 &
  14.6793 &
  12.5160 &
  2.5055 &
  7.5107 &
  0.7126 &
  0.1396 &
  7.1907 \\ \cdashline{2-14} 
\multirow{-7}{*}{\rotatebox[origin=c]{90}{ResNet-50}} &
  MR-MP(SRT) &
  \textbf{3.0833} &
  4.6940 &
  9.4186 &
  3.2926 &
  3.0373 &
  \textbf{3.1649} &
  0.6722 &
  5.3485 &
  3.0103 &
  0.6585 &
  0.0175 &
  9.0671 \\ \hline \hline
 &
  UMR-UMP &
  2.2396 &
  2.7918 &
  5.0826 &
  1.0000 &
  2.7918 &
  1.8959 &
  0.1000 &
  5.0826 &
  2.5913 &
  0.7150 &
  0.0075 &
  11.6725 \\ \cline{2-14} 
 &
  UMR-MP &
  4.6539 &
  8.5698 &
  17.1908 &
  0.9843 &
  8.6056 &
  4.7949 &
  0.0910 &
  17.6305 &
  8.8608 &
  0.4997 &
  0.0121 &
  6.5141 \\ \cdashline{2-14} 
 &
  UMR-MP(T) &
  9.1834 &
  21.4297 &
  35.7315 &
  0.2993 &
  29.0229 &
  14.6611 &
  0.0086 &
  51.6950 &
  25.8518 &
  0.3273 &
  -0.0117 &
  3.7926 \\ \cdashline{2-14} 
 &
  UMR-MP(SRT) &
  \textbf{4.0548} &
  7.1995 &
  14.5421 &
  0.9831 &
  7.2506 &
  \textbf{4.1169} &
  0.0974 &
  14.6495 &
  \textbf{7.3734} &
  0.5295 &
  0.0056 &
  7.2243 \\ \cline{2-14} 
 &
  MR-MP &
  5.0339 &
  9.7305 &
  20.6064 &
  10.1750 &
  3.4003 &
  6.7877 &
  2.6584 &
  6.7137 &
  \textbf{4.6860} &
  0.6624 &
  0.0939 &
  6.6892 \\ \cdashline{2-14} 
 &
  MR-MP(T) &
  8.9175 &
  21.9972 &
  39.7454 &
  99.6336 &
  0.0205 &
  49.8270 &
  96.8945 &
  0.0818 &
  48.4881 &
  0.8281 &
  0.5465 &
  3.9353 \\ \cdashline{2-14} 
\multirow{-7}{*}{\rotatebox[origin=c]{90}{MobileFaceNet}} &
  MR-MP(SRT) &
  \textbf{4.6837} &
  9.0249 &
  18.8782 &
  4.2937 &
  4.9241 &
  \textbf{4.6089} &
  0.9800 &
  9.1016 &
  5.0408 &
  0.6353 &
  0.0301 &
  6.9284 \\ \hline
\end{tabular}%
}
\caption{The achieved verification performance of different experimental settings by ResNet-100, ResNet-50, and MobileFaceNet models along with EUM trained with triplet loss and EUM trained with \gls{srt} loss. The result is reported using synthetically generated masked faces of the IJB-C dataset. The  FMR100\_Th\textsuperscript{UMR-UMP} are equal to 0.1804, 0.2143 and 0.2546  for ResNet-100, ResNet-50 and MobileFaceNet, respectively. The FMR1000\_Th\textsuperscript{UMR-UMP}  are equal to 0.2557, 0.2990  and 0.3493 for ResNet-100, ResNet-50 and MobileFaceNet, respectively.
The lowest EER and the lowest average error of FMR100 and FMR1000 at the defined threshold for each of the evaluation cases and each of the evaluated models are marked in bold.
One can notice the significant improvement in the verification performance induced by our proposed approach (\gls{srt}) in most of the evaluation cases.}
\label{tab:ijbc}
\end{table*}

\section{Conclusion}
\label{sec:conclusion}
In this paper, we presented and evaluated a novel solution to reduce the negative impact of wearing a protective face mask on face recognition performance. 
This work was motivated by the recent evaluation efforts on the effect of masked faces on face recognition performance.
The presented solution is designed to operate on top of existing face recognition models, thus avoiding the need for retraining existing face recognition solutions used for unmasked faces. This goal has been accomplished by proposing the \gls{eum} operated on the embedding space. 
The learning objective of our \gls{eum} is to increase the similarity between genuine unmasked-masked pairs and decrease the similarity between imposter pairs.
We achieved this learning objective by proposing a novel loss function, the \gls{srt} which, unlike triplet loss, dynamically self-adjust its learning objective by concentrating on optimizing the distance between the genuine pairs only when the distance between the imposter pairs is deemed to be sufficient.
Through ablation study and experiments on four masked face datasets and three face recognition models, 
we demonstrated that our proposed \gls{eum} with \gls{srt} significantly improved the masked face verification performance in most experimental settings. 
As a concluding remark, this work is one of the first efforts proposing a solution for masked face recognition without the need for retraining existing face recognition models. 
Several interesting directions of work can be investigated in the future. This includes, but is not limited to studying the demographic effects of masked images on face recognition verification performance and investigating the possibility of dynamically generating realistic masked faces with a variety of mask textures.

\paragraph{\textbf{Acknowledgments}}
This research work has been funded by the German Federal Ministry of Education and Research and the Hessen State Ministry for Higher Education, Research and the Arts within their joint support of the National Research Center for Applied Cybersecurity ATHENE.


{\small
\bibliographystyle{IEEEtran}
\bibliography{main}
}

\end{document}